\documentclass[sigconf]{acmart}

\AtBeginDocument{%
  \providecommand\BibTeX{{%
    Bib\TeX}}}

\copyrightyear{2025}
\acmYear{2025}
\setcopyright{acmlicensed}\acmConference[KDD '25]{Proceedings of the 31st ACM SIGKDD Conference on Knowledge Discovery and Data Mining V.2}{August 3--7, 2025}{Toronto, ON, Canada}
\acmBooktitle{Proceedings of the 31st ACM SIGKDD Conference on Knowledge Discovery and Data Mining V.2 (KDD '25), August 3--7, 2025, Toronto, ON, Canada}
\acmDOI{10.1145/3711896.3737102}
\acmISBN{979-8-4007-1454-2/2025/08}

\acmISBN{978-1-4503-XXXX-X/18/06}

\usepackage{dsfont}
\usepackage{tabularx}
\usepackage{amsmath}
\usepackage{mathtools}
\usepackage{subcaption}
\usepackage{bbm}
\usepackage[vlined,ruled,commentsnumbered]{algorithm2e}
\usepackage{cleveref}
\usepackage{color}
\usepackage{colortbl}
\usepackage{multirow}
\usepackage{array}
\usepackage{listings}
\usepackage{soul}
\usepackage{makecell}
\usepackage{ulem}
\usepackage{enumitem}
\usepackage{todonotes}
\usepackage[most]{tcolorbox}
\usepackage{tikz}
\usepackage{tikz-qtree,tikz-qtree-compat}
\usetikzlibrary{trees,positioning,fit,arrows}
\usepackage[acronym]{glossaries}
\usepackage{glossaries-extra}
\usepackage{mdframed}

\setabbreviationstyle[acronym]{long-short}

\newacronym{lf}{LF}{labeling function}
\newacronym{rbbm}{PWSS}{programmatic weak supervision system}
\newacronym{milp}{MILP}{mixed-integer linear program}

\newacronym{pws}{PWS}{programmatic weak supervision}

\newacronym{ml}{ML}{machine learning}
\newglossaryentry{mlm}{name=model,description={}}
\newglossaryentry{bbm}{name=black box model,description={}}
\newacronym{llm}{LLM}{large language model}
\newglossaryentry{gtl}{name=ground truth label,description={}}
\newglossaryentry{rrp}{name=rule repair problem,description={}}
\newglossaryentry{srrp}{name=single rule repair problem,description={}}

\SetCommentSty{mycommfont}
\tikzset{every tree node/.style={align=center, anchor=north}}
\AtBeginDocument{%
  \providecommand\BibTeX{{%
    \normalfont B\kern-0.5em{\scshape i\kern-0.25em b}\kern-0.8em\TeX}}}

\newbool{Techreport}
\newrobustcmd{\ifnottechreport}[1]{\ifbool{Techreport}{}{#1}}
\newrobustcmd{\iftechreport}[1]{\ifbool{Techreport}{#1}{}}

\newcommand{\paratitle}[1]{\noindent\textbf{{#1}.}}

\usepackage{pythonhighlight}

\newtheorem{theorem}{Theorem}
\newtheorem{proposition}{Proposition}
\newtheorem{example}{Example}

\newtheorem{definition}{Definition}

\newtheorem{lemma}{Lemma}

\newcommand{\proofpar}[1]{\medskip\noindent\underline{#1:}}

\DeclareMathOperator*{\argmax}{argmax}
\DeclareMathOperator*{\argmin}{argmin}

\newcommand{\card}[1]{|\,{#1}\,|}

\newcommand{\indicator}{\mathds{1}\xspace}
\newcommand{\tabhead}[1]{\cellcolor{gray!40}\textbf{#1}}

\newcommand{\oNotation}[1]{O({#1})}
\newcommand{\ptime}{\texttt{PTIME}\xspace}
\newcommand{\npcomplete}{NP-complete\xspace}
\newcommand{\nphard}{NP-hard\xspace}

\newcommand{\oursys}{{\sc RuleCleaner}\xspace}

\newcommand{\model}{\ensuremath{\mathcal{M}}\xspace}
\newcommand{\fullmodel}{\ensuremath{\mathcal{M}_{\rules,\indb}}\xspace}
\newcommand{\indb}{\ensuremath{\mathcal{X}}\xspace}
\newcommand{\x}{\ensuremath{x}\xspace}
\newcommand{\outdb}{\ensuremath{\mathcal{Y}}\xspace}
\newcommand{\lab}{\ensuremath{y}\xspace}
\newcommand{\alabel}[1]{\ensuremath{\text{\upshape \textsf{#1}}}\xspace}

\newcommand{\labOf}[1]{Y[#1]}

\newcommand{\predl}[1]{\ensuremath{\hat{y_{#1}}}\xspace}
\newcommand{\gtl}[1]{\ensuremath{y_{#1}^{*}}\xspace}

\newcommand{\var}{\ensuremath{v}\xspace}

\newcommand{\preds}{\ensuremath{\mathcal{P}}\xspace}

\newcommand{\arulepreds}[1]{\preds_{\rules}}
\newcommand{\pred}{\ensuremath{p}\xspace}

\newcommand{\node}{v}
\newcommand{\apath}{\ensuremath{P}\xspace}

\newcommand{\leafpaths}[1]{\ensuremath{leafpaths(#1)}\xspace}
\newcommand{\plast}[1]{last(#1)}

\newcommand{\rpath}[2]{\ensuremath{P[#1,#2]}\xspace}

\newcommand{\assigns}{\ensuremath{\Lambda}\xspace}
\newcommand{\cdp}{\ensuremath{\mathcal{X}}\xspace}

\newcommand{\pathdps}[1]{\ensuremath{\mathcal{X}_{#1}}\xspace}

\newcommand{\rl}{\ensuremath{r}\xspace}

\newcommand{\rules}{\ensuremath{\mathcal{R}}\xspace}

\newcommand{\repairrules}{\ensuremath{{\rules^{*}}}\xspace}

\newcommand{\ltrue}{\ensuremath{\text{\upshape\textbf{true}}}\xspace}
\newcommand{\lfalse}{\ensuremath{\text{\upshape\textbf{false}}}\xspace}

\newcommand{\rresult}[2]{\ensuremath{#1(#2)}\xspace}
\newcommand{\reval}[2]{\ensuremath{\text{\upshape \textsc{eval}}(#1,#2)}\xspace}
\newcommand{\areval}{\reval{\rnode}{\x}}
\newcommand{\rooteval}[2]{\reval{\rroot{#1}}{#2}}
\newcommand{\rroot}[1]{\ensuremath{\textsc{root}(#1)}\xspace}
\newcommand{\rnode}{\ensuremath{n}\xspace}
\newcommand{\tchild}[1]{\ensuremath{C_{\ltrue}(#1)}\xspace}
\newcommand{\fchild}[1]{\ensuremath{C_{\lfalse}(#1)}\xspace}

\newcommand{\rrefine}[5]{\text{\upshape\textbf{refine}}(#1,#2,#3,#4,#5)}
\newcommand{\rrefinelabel}[3]{\text{\upshape\textbf{refine}}(#1,#2,#3)}

\newcommand{\complaints}{\ensuremath{\mathcal{Z}}\xspace}
\newcommand{\xcomplaints}{\ensuremath{\indb^{*}}\xspace}

\newcommand{\rop}{\ensuremath{\phi}\xspace}
\newcommand{\rseq}{\ensuremath{\Phi}\xspace}

\newcommand{\rcost}{cost}
\newcommand{\pcost}{rcost}

\newcommand{\aparam}[1]{\tau_{#1}}

\newcommand{\iat}{\ensuremath{\aparam{acc}\xspace}}
\newcommand{\inat}{\ensuremath{\aparam{E}\xspace}}
\newcommand{\lat}{\ensuremath{\aparam{racc}\xspace}}

\newcommand{\fcall}[1]{\ensuremath{\text{\texttt{#1}}}\xspace}
\newcommand{\fgetsepp}{\fcall{GetSeperatorPred}}
\newcommand{\fgetcovpred}{\fcall{GetCoveringPred}}
\newcommand{\frepairrule}{\fcall{SingleRuleRefine}}
\newcommand{\fgetallpreds}{\fcall{GetAllCandPredicates}}

\newcommand{\fappend}{\fcall{append}}
\newcommand{\fpathrefine}{\fcall{RefinePath}}

\newcommand{\atomicunit}{\ensuremath{\mathcal{A}}\xspace}

\newcommand{\au}{\ensuremath{a}\xspace}
\newcommand{\ac}[2]{#1[#2]}

\newcommand{\defaultlabel}{{\ensuremath{y_0}\xspace}}

\newcommand{\evidence}{\textsc{\upshape Evidence}\xspace}
\newcommand{\iaccuracy}{\textsc{\upshape Acc}\xspace}
\newcommand{\laccuracy}{\textsc{\upshape Acc}\xspace}

\newcommand{\positive}{\alabel{POS}}
\newcommand{\negative}{\alabel{NEG}}
\newcommand{\abstain}{\alabel{ABSTAIN}}

\newcommand{\pdps}[1]{\ensuremath{\mathcal{X}_{#1}}\xspace}
\newcommand{\pz}[1]{\ensuremath{\mathcal{Z}_{#1}}\xspace}
\newcommand{\pathc}[1]{\ensuremath{\mathcal{Z}_{#1}}\xspace}
\newcommand{\fixp}{\ensuremath{P_{fix}}\xspace}
\newcommand{\plabels}{\ensuremath{Y}\xspace}
\newcommand{\asslabel}{\ensuremath{Z}\xspace}

\newcommand{\gini}{\ensuremath{I_G}\xspace}
\newcommand{\gi}{\ensuremath{I_G}\xspace}

\newcommand{\cut}[1]{}

\newcommand{\abbrRBBM}{\gls{rbbm}\xspace}
\newcommand{\abbrRBBMs}{\glspl{rbbm}\xspace}
\newcommand{\abbrGreedyPredRepair}{GreedyPathRepair\xspace}
\newcommand{\abbrEntropyPathRepair}{EntropyPathRepair\xspace}
\newcommand{\abbrOptimalPathRepair}{BruteForcePathRepair\xspace}

\newcommand{\eat}[1]{}

\newcommand{\red}[1]{\textcolor{red}{#1}}

\newcommand{\blue}[1]{\textcolor{blue}{#1}}

\definecolor{white}{rgb}{1,1,1}
\definecolor{black}{rgb}{0,0,0}
\definecolor{grey}{rgb}{0.7,0.7,0.7}
\definecolor{dgrey}{rgb}{0.5,0.5,0.5}
\definecolor{lightgrey}{rgb}{0.88,0.88,0.88}
\definecolor{lgrey}{rgb}{0.9,0.9,0.9}
\definecolor{llgrey}{rgb}{0.93,0.93,0.93}
\definecolor{lllgrey}{rgb}{0.96,0.96,0.96}
\definecolor{tableHeadGray}{rgb}{0.85,0.85,0.85}
\definecolor{oddRowGrey}{rgb}{0.95,0.95,0.95}
\definecolor{evenRowGrey}{rgb}{0.85,0.85,0.85}
\definecolor{yellow}{rgb}{1.0, 1.0, 0.0}
\definecolor{lightyellow}{rgb}{1.0, 1.0, 0.88}
\definecolor{selectiveyellow}{rgb}{1.0, 0.73, 0.0}
\definecolor{shadered}{rgb}{1,0.85,0.85}
\definecolor{red}{rgb}{1,0,0}
\definecolor{shadegreen}{rgb}{0.95,1,0.95}
\definecolor{green}{rgb}{0,1,0}
\definecolor{darkgreen}{rgb}{0,0.5,0}
\definecolor{shadeblue}{rgb}{0.95,0.95,1}
\definecolor{blue}{rgb}{0,0,1}
\definecolor{darkblue}{rgb}{0,0,0.5}
\definecolor{darkpurple}{rgb}{0.5,0,0.5}
\definecolor{darkdarkpurple}{rgb}{0.3,0,0.3}
\definecolor{toplogiccolor}{rgb}{1,1,1}
\definecolor{bottomlogiccolor}{rgb}{1,1,1}

\Crefname{section}{Section}{Section}
\Crefname{figure}{Figure}{Figure}
\Crefname{table}{Table}{Table}
\Crefname{equation}{Eq.}{Eq.}
\Crefname{lemma}{Lemma}{Lemma}
\Crefname{theorem}{Theorem}{Theorem}
\Crefname{definition}{Definition}{Definition}
\Crefname{example}{Example}{Example}
\Crefname{proposition}{Proposition}{Proposition}
\Crefname{corollary}{Corollary}{Corollary}
\Crefname{appendix}{Appendix}{Appendix}

\crefname{section}{sec.}{sec.}
\crefname{figure}{fig.}{fig.}
\crefname{table}{tab.}{tab.}
\crefname{equation}{eq.}{eq.}
\crefname{lemma}{lem.}{lem.}
\crefname{theorem}{thm.}{thm.}
\crefname{definition}{def.}{def.}
\crefname{example}{ex.}{ex.}
\crefname{proposition}{prop.}{prop.}
\crefname{corollary}{cor.}{cor.}

\acmSubmissionID{rtfb1871}

\begin{document}

\title{Refining Labeling Functions with Limited Labeled Data}

\author{Chenjie Li}
\affiliation{%
  \institution{University of Illinois Chicago}
  \city{Chicago}
  \state{IL}
  \postcode{60607}
  \country{USA}
}
\email{cl206@uic.edu}

\author{Amir Gilad}
\orcid{[OPTIONAL-ORCID]}
\affiliation{%
  \institution{Hebrew University}
  \city{Jerusalem}
  \country{Israel}
}
\email{amirg@cs.huji.ac.il}

\author{Boris Glavic}
\affiliation{%
  \institution{University of Illinois Chicago}
  \city{Chicago}
  \state{IL}
  \postcode{60607}
  \country{USA}
}
\email{bglavic@uic.edu}

\author{Zhengjie Miao}
\affiliation{%
  \institution{Simon Fraser University}
  \city{Burnaby}
  \state{BC}
  \postcode{V5A 1S6}
  \country{Canada}
}
\email{zhengjie@sfu.ca}

\author{Sudeepa Roy}
\orcid{[OPTIONAL-ORCID]}
\affiliation{%
  \institution{Duke University}
  \city{Durham}
  \state{NC}
  \postcode{27708}
  \country{USA}
}
\email{sudeepa@cs.duke.edu}


\begin{abstract}
 Programmatic weak supervision (PWS) 
 significantly reduces human effort for labeling data by combining the outputs of user-provided \glspl{lf} on unlabeled datapoints. 
   However, 
   the quality of the generated labels depends directly on the accuracy of the \glspl{lf}.
  In this work, we study the problem of fixing \glspl{lf} based on a small set of labeled examples. 
Towards this goal, we develop novel techniques for repairing a set of \glspl{lf} by minimally changing their results on the labeled examples such that the fixed \glspl{lf} ensure that (i) there is sufficient evidence for the correct label of each labeled datapoint and (ii) the accuracy of each repaired LF is sufficiently high.
We model \glspl{lf} as conditional rules, which enables us to refine them, i.e., to selectively change their output for some inputs. We demonstrate experimentally that our system improves the quality of \glspl{lf} based on surprisingly small sets of labeled datapoints.
\end{abstract}


\tikzset{
  level 1/.style={level distance=1.5cm},  
  level 2/.style={level distance=1.5cm},  
  level 3/.style={level distance=1.5cm},  
  level 4/.style={level distance=2cm},   
  assnode/.style = {shape=rectangle, rounded corners, align=center},
  snode/.style = {shape=rectangle, rounded corners, draw, align=center, top color=white, bottom color=red!20},
  newsnode/.style = {shape=rectangle, rounded corners, draw, align=center, top color=white, bottom color=green!20},
  newlogic/.style = {shape=rectangle, rounded corners, draw, align=center, top color=white, bottom color=green!20},->,
  logic/.style = {shape=rectangle, rounded corners, draw, align=center, top color=toplogiccolor, bottom color=bottomlogiccolor},-
}

\normalem



\maketitle

\newcommand\kddavailabilityurl{https://doi.org/10.5281/zenodo.15558280}

\ifdefempty{\kddavailabilityurl}{}{
\begingroup\small\noindent\raggedright\textbf{KDD Availability Link:}\\
The source code of this paper has been made publicly available at \url{\kddavailabilityurl}.
\endgroup
}






\setbool{Techreport}{true}


\begin{figure}[t]
  \centering
  \begin{minipage}{1\linewidth}
    \centering
    \begin{python}
def key_word_star(v): #LF-1
  words = ['star', 'stars']
  return POS if words.intersection(v) else ABSTAIN

def key_word_waste(v): #LF-2
   return NEG if ('waste' in v) else ABSTAIN

def key_word_poor(v): #LF-3
  words = ['poorly', 'useless', 'horrible', 'money']
  return NEG if words.intersection(v) else ABSTAIN
\end{python}
    \subcaption{Three example \glspl{lf} for amazon reviews} 
    \label{fig:example_lfs_witan}
  \end{minipage}
  \begin{minipage}{0.49\linewidth}
    \centering\footnotesize
    \begin{tikzpicture}
      [
      scale=.58,
      sibling distance=3cm
      ]
      \node[logic](a) {\pyth{'star' or 'stars' in v}}
          child {node[snode](b) {$\abstain$}
            edge from parent node[rectangle,left] {\small\lfalse}}
          child {
            node[newsnode](c) {\pyth{'one' in v}}
            child {node[newsnode](d) {$\positive$}
              edge from parent node[rectangle, left] {\small\lfalse}}
            child {node[newsnode](e) {$\negative$}
              edge from parent node[rectangle,right] {\small\ltrue}}
            edge from parent node[rectangle, right] {\small\ltrue}
        };
\end{tikzpicture}
    \subcaption{Refined rule for LF-1} \label{fig:example_lf_stars}
  \end{minipage}
  \begin{minipage}{0.49\linewidth}
    \centering\footnotesize

    \begin{tikzpicture}[scale=.58, sibling distance=30mm]
      \node[logic](a) {\pyth{words.intersection(v)}}
        child {node[snode](b) {$\abstain$}
          edge from parent node[rectangle, left] {\small\lfalse}}
        child {
          node[newlogic](c) {\pyth{'yes' in v}}
          child {node[newsnode](d) {$\negative$}
            edge from parent node[rectangle, left] {\small\lfalse}}
          child {node[newsnode](e) {$\positive$}
            edge from parent node[rectangle, right] {\small\ltrue}}
          edge from parent node[rectangle, right] {\small\ltrue}
      };
    \end{tikzpicture}
    \subcaption{Refined rule for LF-3} \label{fig:example_lf_poor_words}
  \end{minipage}\\[-3mm]
  \caption{\glspl{lf} before / after refinement by \oursys}
  \label{fig:rules-old-new}
\end{figure}



\begin{table*}[htbp]
\footnotesize
    \centering
    \begin{tabular}{|c|>{\centering}m{0.43\linewidth}|>{\centering}m{6mm}|>{\centering}m{20mm}|>{\centering}m{20mm}|c|c|c|}
        \hline
        \cellcolor{gray!40} & \cellcolor{gray!40} & \cellcolor{gray!40} & \cellcolor{gray!40} & \cellcolor{gray!40} & \multicolumn{3}{c|}{\cellcolor{gray!40}\textbf{LF labels: old (new)}} \\
        \cellcolor{gray!40} \multirow[c]{-2}{4mm}{\cellcolor{gray!40}\textbf{id}}
                            & \centering \multirow[c]{-2}{=}{\cellcolor{gray!40}\centering\textbf{text}}
                                                  & \multirow[c]{-2}{=}{\cellcolor{gray!40}\centering\textbf{true label}}
                                                                        & \multirow[c]{-2}{=}{\cellcolor{gray!40}\centering\textbf{old predicted  label by Snorkel}}
                                                                                              & \multirow[c]{-2}{=}{\cellcolor{gray!40}\centering\textbf{new predicted label by Snorkel }}
                                                                                                                    & \cellcolor{gray!40}\textbf{1} & \cellcolor{gray!40}\textbf{2} & \cellcolor{gray!40}\textbf{3}  \\
        \hline
        0 & five stars. product works fine & P & P & P & P & - & - \\
        \hline
        1 & one star. rather poorly written needs more content and an editor & N & \red{P} & \blue{N} & \sout{P}~(\textcolor{blue}{N}) & - & N \\
        \hline
        2 & five stars. awesome for the price lightweight and sturdy & P & P & P & P & - & - \\
        \hline
        3 & one star. not my subject of interest, too dark & N & \red{P} & \blue{N} & {\sout{P}}~(\textcolor{blue}{N}) & - & - \\
        \hline
        4 & yes, get it! the best money on a pool that we have ever spent. really cute and holds up well with kids constantly playing in it & P & \red{N} & \blue{P} & - & - & {\sout{N}}~(\textcolor{blue}{P})  \\
        \hline
      \end{tabular}
\vspace{1mm}
  \caption{\label{exp:intro_youtube_table} Products reviews with ground truth labels ("P"ositive or "N" egative), predicted labels by Snorkel~\cite{RatnerBEFWR20} (before and after rule repair), and the results of the \glspl{lf} from \Cref{fig:rules-old-new} ("-" means \abstain). Results for repaired rules are shown in blue.}
  \vspace{-5mm}
  \label{fig:dataset-amazon}
\end{table*}

\section{Introduction}\label{sec:introduction}
\Gls{pws}~\cite{RatnerBEFWR20, zhang-22-spws} is a powerful 
technique for creating training data. Unlike manual labeling, where labels are painstakingly assigned by hand to each training datapoint, data programming assigns labels by combining the outputs of \glsentryfullpl{lf} --- heuristics that take a datapoint as input and output a label --- using a \gls{mlm}. This approach dramatically reduces the human effort required to label data. To push this reduction even further, recent approaches automate the generation of \glspl{lf}~\cite{DenhamLSN22,boecking2021interactive,VarmaR18, DBLP:conf/edbt/GuanCK25}. For example, Witan~\cite{DenhamLSN22} creates \glspl{lf} from simple predicates that are effective in differentiating datapoints, subsequently guiding users to select and refine sensible \glspl{lf}. Guan et al.~\cite{DBLP:conf/edbt/GuanCK25} employ \glspl{llm} to derive \glspl{lf} based on a small amount of labeled data, further reducing the dependency on human intervention. One advantage of \gls{pws} over weak supervision with a \gls{bbm} is that \glspl{lf} are inherently interpretable.

Regardless of whether \glspl{lf} are manually crafted by domain experts or generated by automated techniques, users face significant challenges when it comes to 
repairing these \glspl{lf} to correct issues with the resulting labeled data. 
The black-box nature of the model that combines \glspl{lf} results obscures which specific \glspl{lf} are responsible for mislabeling a datapoint, and large training datasets make it difficult for users to 
manually identify effective repairs. 
While explanation techniques for \gls{pws}~\cite{zhang-22-unpwssinf, guan-24-w} can identify \glspl{lf} responsible for erroneous labels, determining how to repair the \glspl{lf} to fix these errors remains a significant challenge.

{\em In this work, we tackle the challenge of automatically suggesting repairs for a set of \glspl{lf} based on a small set of labeled datapoints.} Our approach 
refines an \gls{lf} by locally overriding its outputs to align with expectations for specific datapoints. Rather than replacing human domain expertise or existing automated \gls{lf} generation, our method, \oursys,  
 improves an existing set of \glspl{lf}. Our approach is versatile, 
supporting arbitrarily complex \glspl{lf}, and various \glspl{bbm} that combine them such as Snorkel~\cite{RatnerBEFWR20}, FlyingSquid~\cite{fu2020fast} or simpler models like majority voting. \oursys is agnostic to the source of \glspl{lf}, enabling the repair of \glspl{lf} generated by tools like Witan~\cite{DenhamLSN22}, \glspl{llm}~\cite{DBLP:conf/edbt/GuanCK25}, and those created by domain experts.

To address the challenge of refining \glspl{lf} expressed in a general-purpose programming language, we model \glspl{lf} as {\em rules}, represented as trees. In these trees, inner nodes are {\em predicates},  Boolean conditions evaluated on datapoints, 
and leaf nodes correspond to labels. Such a tree encodes a cascading series of conditions, starting at the root, each predicate directs navigation to a {\em true} or {\em false} child until a leaf node is reached, which assigns the label to the input datapoint. \emph{This model can represent any \gls{lf} as a rule} by creating predicates that match the result of the \gls{lf} against every possible label (see \ifnottechreport{\cite{li2025refininglabelingfunctionslimited}}\iftechreport{App.~\ref{sec:rule-conversion}}).



\begin{example}\label{ex:lf_intro_witan} Consider the \textit{Amazon Review Dataset} from~\cite{DBLP:conf/www/HeM16, DenhamLSN22} which contains reviews for products bought from Amazon and the task of labeling the reviews as  \positive or P (positive), or \negative or N (negative). 
A subset of \glspl{lf} generated by the Witan system~\cite{DenhamLSN22} for this task are shown in \Cref{fig:example_lfs_witan}. \texttt{key\_word\_star} labels reviews as \positive that contain either \emph{``star''} or \emph{``stars''} and otherwise returns \abstain 
(the function cannot make a prediction).
Some reviews with their ground truth labels (unknown to the user)
and the labels predicted by \emph{Snorkel}~\cite{RatnerSWSR16} are shown in \Cref{fig:dataset-amazon}, which
also shows the three \glspl{lf} 
from \Cref{fig:example_lfs_witan}. Reviews 1, 3, and 4 are mispredicted by the model trained by Snorkel over the \gls{lf} outputs. 
 Our goal is to reduce such misclassifications by refining the \glspl{lf}. We treat systems like Snorkel as a blackbox that can use any algorithm or \gls{mlm} to generate labels. 

%
Suppose that for a small set of reviews, the true label is known (\Cref{fig:dataset-amazon}).  
\oursys uses these \glspl{gtl}
to generate a set of repairs 
for the \glspl{lf} 
by refining \glspl{lf} to align them with the ground truth. \Cref{fig:dataset-amazon} also shows the labels produced by the repaired \glspl{lf} 
(updated labels shown in blue), 
 and the predictions generated by 
Snorkel before and after \gls{lf} repair.
 \oursys repairs LF-1 and LF-3 from \Cref{fig:example_lfs_witan} by adding new predicates (refinement).  \Cref{fig:example_lf_stars,fig:example_lf_poor_words} show the refined rules
 in tree form 
 with new nodes highlighted in green.
Consider the repair for LF-1.
Intuitively, this repair is sensible: a review mentioning \emph{``one''} and \emph{``star(s)''} is likely negative. 
\end{example}

Our \oursys system produces repairs as shown in the example above. We make the following contributions.


\begin{itemize}[leftmargin=*]
\item {\bf The \gls{pws} Repair Problem}. We introduce a general model for \gls{pws} as
\glspl{rbbm}, where interpretable \glspl{lf} (rules) are combined to predict labels for a dataset \indb (\textbf{\Cref{sec:model}}). 
  We formalize the problem of repairing \glspl{lf} in \abbrRBBM  through refinement, proving the problem to be \nphard. To avoid overfitting, we (i) minimize the changes to the outputs of the original \glspl{lf} and (ii) allow some \glspl{lf} to return incorrect labels for some labeled datapoints.

\item {\bf Efficient Rule Repair Algorithm}. 
In spite of its hardness,
in practice it is feasible to solve the repair problem exactly as the number of labeled examples is typically small. We formalize this problem as a \gls{milp} that determines changes to the \gls{lf} output on the labeled examples (\textbf{\Cref{sec:repair}}). To implement these changes, we refine individual rules to match the desired outputs (\textbf{\Cref{sec:single-rule-repairs}}). To further decrease the likelihood of overfitting and limit the complexity of the fixed rules, we want to minimize the number of new predicates that are added. This problem is also \nphard. We propose a \ptime information-theoretic heuristic algorithm (\textbf{\Cref{sec:rule-refin-repa}}).

\item {\bf Comprehensive Experimental Evaluation}.
We conduct experiments on 11 real datasets using Snorkel~\cite{RatnerBEFWR20} over \glspl{lf} generated by Witan~\cite{DenhamLSN22} or \glspl{llm}~\cite{DBLP:conf/edbt/GuanCK25} (\textbf{\Cref{sec:experiments}}). Furthermore, we compare against using \glspl{llm} directly for labeling and for repairing \glspl{lf}.
\oursys significantly improves labeling accuracy using a small number of labeled examples. While direct labeling with \glspl{llm} achieves impressive accuracy for advanced models like GPT-4o, it is also prohibitively expensive.
\end{itemize}


\section{The \oursys\ Framework}\label{sec:model}

As shown in \Cref{fig:framework}, we assume as input a set of \glspl{lf} modeled as rules $\rules$, the corresponding labels produced by an \abbrRBBM for an unlabeled dataset $\indb$, and a small subset of labeled datapoints $\xcomplaints \subset \indb$.
 \oursys refines specific \glspl{lf} based on this input, generating an updated set of rules $\repairrules$. Finally, the \abbrRBBM applies these revised rules to re-label the dataset.

\begin{figure}[t]
    \centering
    \includegraphics[scale=0.25]{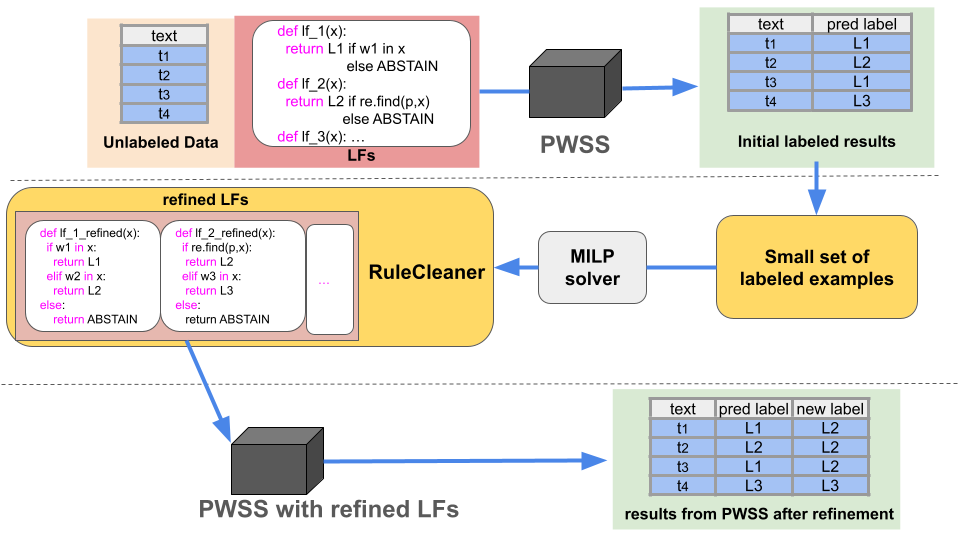}
    \caption{The \oursys\ framework for repairing \glspl{lf} (rules \rules). After running an \abbrRBBM (e.g., Snorkel) on the rule outputs, \oursys fixes the rules $\rules$ using labeled examples $\xcomplaints$. Finally, the \abbrRBBM is rerun on the output of the refined rules $\repairrules$ on the whole dataset $\indb$ to produce the repaired labels.}
    \label{fig:framework}
\end{figure}

\subsection{Rules and \Glspl{rbbm}}\label{sec:data-and-model}

To be able to repair a \gls{lf} by selectively overriding its output based on conditions that hold for an input datapoint, we model \glspl{lf} as a set of cascading conditions. A rule \rl is a \emph{tree} where leaf nodes represent labels from a set of labels \outdb and the non-leaf nodes are labeled with Boolean predicates from a space of predicates \preds. Each non-leaf node has two outgoing edges labeled with \ltrue and \lfalse. A rule \rl assigns a label $\rl(\x)$ to an input datapoint \x by evaluating the predicate at the root, following the outgoing edge \ltrue if the predicate evaluates to true and the \lfalse edge otherwise. Then the predicate of the node at the end of the edge is evaluated. This process is repeated until a leaf node is reached. The label of the leaf node is the label assigned by \rl to \x.
\begin{example}
\Cref{fig:lf-dc-to-rule}
shows the rule for a \gls{lf} that returns \negative if the
review contains the word `{\em waste}' and returns
$\abstain$ otherwise.
In \iftechreport{\Cref{prop:rule-translation-cor}}\ifnottechreport{\cite{li2025refininglabelingfunctionslimited}}, we show how to translate any \gls{lf} written in a general-purpose programming language into a rule in \ptime.
\end{example}
\ifnottechreport{
    To demonstrate that \oursys can support arbitrary LFs, including those with complex components, consider an LF that first checks whether the input text contains the keywords \texttt{`star'} or \texttt{`stars'}. If so, it returns the label POSITIVE. Otherwise, it calls an external sentiment analysis function, returning POSITIVE only if the sentiment score exceeds 0.7, and ABSTAIN otherwise. The first condition is directly translated into a predicate in the rule tree, while the sentiment analysis branch is wrapped as a black-box component \pyth{blackbox\_lf}. The translated tree rule is shown in \Cref{fig:translate-partial-blackbox}. Any black-box LF can be translated into a rule tree using our Translate-BBox algorithm~\cite{li2025refininglabelingfunctionslimited}. From here on, we will use the terms \gls{lf} and rule interchangeably.

 \begin{figure}[t]
    \centering
    \begin{tikzpicture}[scale=.8,
    level 1/.style={level distance=1.5cm, sibling distance=3cm},
    level 2/.style={level distance=1.5cm,sibling distance=3cm},
    level 3/.style={level distance=1.5cm, sibling distance=3cm},
    level 4/.style={level distance=1.5cm, sibling distance=2cm},
     level 5/.style={level distance=1.5cm, sibling distance=2cm}]

    \node[logic] {\pyth{['star', 'stars'].intersection(v)}}
        child {
        node[logic]{\pyth{blackbox\_lf}(\var) = \positive}
          child {
          node[logic]{\pyth{blackbox\_lf}(\var) = \negative}
            child {
            node[snode]{$\abstain$} edge from parent node[above] {{\small\lfalse}}
            }
            child {node[snode] {$\negative$} edge from parent node[above] {{\small\ltrue}}}
            edge from parent node[above]  {{\small\lfalse}}
          }
          child {node[snode] {$\positive$} edge from parent node[above]  {{\small\ltrue}}}
          edge from parent node[above]  {{\small\lfalse}}
        }
        child {node[snode] {$\positive$} edge from parent node[above]  {{\small\ltrue}}};
    \end{tikzpicture}
    \caption{Translating a LF wrapping parts into a blackbox function} \label{fig:translate-partial-blackbox}
\end{figure}
}
%

\begin{figure}[t]
  \centering
  \begin{minipage}{0.39\linewidth}
    \begin{tikzpicture}[scale=.75,
    level 1/.style={level distance=1.3cm},  
      sibling distance=2cm
  ]
      \node[logic](a) {\pyth{'waste' in v}}
          child {node[snode](b) {$\abstain$}
            edge from parent node[rectangle,left, near start] {\small\lfalse}}
          child {node[snode](c) {$\negative$}
            edge from parent node[rectangle,right, near start] {\small\ltrue}};

    \end{tikzpicture}
  \end{minipage}
  \caption {Rule form  of the LF \pyth{keyword_word_waste} (\Cref{fig:example_lfs_witan})}
   \label{fig:lf-dc-to-rule}
\end{figure}
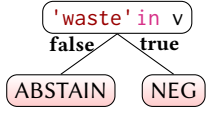

Consider a set of input {\em datapoints} $\indb$ and a set of discrete {\em labels} $\outdb$.  For a datapoint $\x \in \indb$,
 $\gtl{\x}$ denotes the datapoint's (unknown) true label.
 A \abbrRBBM takes \indb, the labels \outdb, and a set of rules \rules  as input and produces a {\em model} $\fullmodel$ as the output that maps each datapoint in  $\indb$ to a label in $\outdb$.
Without loss of generality, we assume the presence of an \emph{abstain label} $\defaultlabel \in \outdb$ that is used by the \abbrRBBM or a rule to abstain from providing a label to some datapoints.

\begin{definition}[\gls{rbbm}]\label{def:black-box}
Given a set of datapoints \indb, a set of labels \outdb, and a set of rules \rules, a
\abbrRBBM takes $\rules(\indb)$ as input and produces a model  $\fullmodel: \indb \to \outdb$ that maps datapoints $\x \in \indb$  to labels

\[
  \fullmodel(\x) = \predl{\x}.
\]

\end{definition}
In the following, we will often drop $\rules$ and $\indb$ from $\fullmodel$ when they are irrelevant to the discussion.


\subsection{The Rule Repair Problem}\label{sec:rule-repair-problem}


\paratitle{Rule Refinement Repairs}
We model a repair of a set of rules $\rules$ as a {\bf repair sequence} $\rseq = \rop_1, \ldots, \rop_k$ of refinement steps $\rop_i$ and use $\repairrules = \{\rl_1', \ldots, \rl_m'\}$ to denote $\rseq(\rules)$.
We repair rules by
\textit{refining} them by replacing a leaf node with a new predicate to achieve a desired change to the rule's result on some datapoints.
  Consider a rule $\rl$, a path $\apath$ ending in a node $\rnode$, and a predicate $\pred$ and two labels $\lab_1$ and $\lab_2$.
The refinement $\rrefine{\rl}{\apath}{\pred}{\lab_1}{\lab_2}$ of $\rl$ replaces $\rnode$ with a new node labeled $\pred$ and adds the new leaf nodes for $\lab_1, \lab_2$:
\[
  \rl\left[\rnode \gets     \begin{tikzpicture}[baseline=-12pt,scale=.7,
      level 1/.style={level distance=1.2cm, sibling distance=2cm},
      snode/.style = {shape=rectangle, rounded corners, draw, align=center, top color=white, bottom color=red!20},
      newsnode/.style = {shape=rectangle, rounded corners, draw, align=center, top color=white, bottom color=green!20},
      logic/.style = {shape=rectangle, rounded corners, draw, align=center, top color=toplogiccolor, bottom color=bottomlogiccolor},
      newlogic/.style = {shape=rectangle, rounded corners, draw, align=center, top color=white, bottom color=green!20},-]

      \node[logic](a) {$p$}
        child {node[logic](b) {$\lab_1$}
            edge from parent node[rectangle,left, near start] {{\small\lfalse}}}
        child {node[logic](c) {$\lab_2$}
            edge from parent node[rectangle,right, near start] {{\small\ltrue}}};

    \end{tikzpicture}\right]
\]

For example, \Cref{fig:example_lf_stars} shows the result of refinement where a leaf $\positive$ was replaced with the subtree highlighted in green. 

\paratitle{Desiderata}
Given the labeled training data $\xcomplaints$, we would like the repaired rules to provide sufficient information about the true labels for datapoints in $\xcomplaints$ to the \abbrRBBM without overfitting to the small number of labeled datapoints in $\xcomplaints$. Specifically, we want the repair to fulfill the following desiderata:

\paratitle{Datapoint Evidence}
 We define the \emph{evidence} for a datapoint $\x_i$ as the fraction of non-abstain labels ($\neq$\defaultlabel) the datapoint receives from the $m$ rules in \rules. The repaired rules should provide sufficient evidence for each datapoint $\x_i$, such that the \abbrRBBM can make an informed decision about $\x_i$'s label.

\[
\evidence(\x_i) =\frac{\sum_j \indicator[\rl_j'(\x_i) \neq \defaultlabel]}{m}
\]

\paratitle{Datapoint Accuracy}
The \emph{accuracy} for a datapoint $\x_i$ is defined below. The accuracy of the repaired rules that do not abstain on $\x_i$ should be high.
\[
  \iaccuracy(\x_i) = \frac{\sum_{j:\,\rl_j'(\x_i) \neq \defaultlabel} \indicator[\rl_j'(\x_i) = \gtl{\x_i}]}{m}
\]

\paratitle{Rule Accuracy}
In addition, the rules should have high accuracy. The \emph{accuracy} of a rule $\rl_j \in \repairrules$ is defined as the fraction of the $n$ datapoints in $\xcomplaints$ on which it returns the ground truth label.
\[
\laccuracy(\rl_j') = \frac{\sum_{i:\,\rl_j'(\x_i) \neq \defaultlabel} \indicator[\rl_j'(\x_i) = \gtl{\x_i}]}{n}
\]

\paratitle{Repair Cost}
For a repair sequence $\rseq$ and $\repairrules = \rseq(\rules)$ we define its cost as the number of labels that differ between the results of $\rules = \{\rl_1, \ldots, \rl_m\}$ and $\repairrules = \{\rl_1', \ldots, \rl_m'\}$ on $\xcomplaints$. Optimizing for low repair cost avoids overfitting to \xcomplaints and preserves rule semantics where feasible.
\[
\rcost(\rseq) = \sum \indicator[\rl_j(\x_i) \neq \rl_j'(\x_i)]
\]

We state the rule repair problem as an optimization problem: minimize the number of changes to labeling function results ($\rcost(\rseq)$) while ensuring the desiderata enforced by thresholds $\inat$ (evidence), $\iat$ (accuracy), and $\lat$ (rule accuracy).

\begin{definition}[\Gls{rrp}]\label{def:repair-problem-old}
Consider a black-box model $\fullmodel$ that uses a set of $m$ rules $\rules$,  a dataset of $n$ datapoints $\indb$,  output labels $\outdb$, and ground truth labels for a subset of datapoints $\xcomplaints$.
Given thresholds $\iat \in [0,1]$, $\inat \in [0,1]$, and $\lat \in [0,1]$,
the \emph{\gls{rrp}} is to find a repair sequence $\rseq$ such that:
%
\begin{align*}
 {\text{\bf argmin}}_{\,\rseq} &\hspace{3mm} \rcost(\rseq)\\ 
 \text{\bf subject to}
 &\hspace{3mm}\forall\, i \in [1,n]: \iaccuracy(\x_i) \geq \iat \land
 \evidence(\x_i) \geq \inat\\
 &\hspace{3mm}\forall\, j \in [1,m]: \laccuracy(\rl_j') \geq \lat\\
\end{align*}
\end{definition}
Note that since we treat the \abbrRBBM as a black box, we can, in general, not guarantee that the \abbrRBBM's performance on the unlabeled dataset $\indb$ will improve. Nonetheless, we will demonstrate experimentally in \Cref{sec:experiments} that significant improvements in the accuracy of rules on $\indb$ can be achieved based on 10s of training examples. This is due to the use of predicates in rule repairs that generalize beyond \xcomplaints. 
While finding an optimal repair is \nphard, we can still solve this problem exactly as $\xcomplaints$ is expected to be small.

\begin{theorem}\label{theo:rule-repair-hardness}
The rule repair problem is \nphard in the size of $\rules$.
\end{theorem}


\section{Ruleset Repair Algorithm}\label{sec:repair}
We now present an algorithm that solves the \gls{rrp} in two steps. In the first step, we use an \gls{milp} to determine desired changes to the outputs of rules, and in the second step, described in \Cref{sec:single-rule-repairs}, we implement these changes by refining individual rules to return the desired output on \xcomplaints. 

\subsection{MILP Formulation}
\label{sec:milp-formulation}

\newcommand{\newresultvar}{\ensuremath{o}\xspace}
\newcommand{\changedvar}{\ensuremath{m}\xspace}
\newcommand{\correctvar}{\ensuremath{c}\xspace}
\newcommand{\abstainvar}{\ensuremath{e}\xspace}

In the \gls{milp}, we use an integer variable $\newresultvar_{ij}$ for each datapoint $\x_i \in \xcomplaints$ and rule $\rl_j$ that stores the label that the repaired rule $\rl_j'$ should assign to $\x_i$. That is, in combination these variables store the desired changes to the results of rules that we then have to implement by refining each rule $\rl_j$ to a rule $\rl_j'$.  
We restrict these variables to take values in $[0, \card{\outdb} - 1]$ where value $i$ represents the label $\lab_i \in \outdb$ with $0$ encoding \defaultlabel. 
To encode the objective (minimizing the changes to the outputs of rules on \xcomplaints) we use a Boolean variable $\changedvar_{ij}$ for each rule $\rl_j$ and datapoint $\x_i$ that is $1$ iff $\newresultvar_{ij} \neq \rl_j(\x_i)$ (the output of $\rl_j'$ on $\x_i$ is different from $\rl_j(\x_i)$). The objective is then to minimize the sum of these indicators $\changedvar_{ij}$.

To encode the side constraints of the \gls{rrp}, we introduce additional indicators: $\correctvar_{ij}$ is $1$ if $\newresultvar_{ij}= \gtl{\x_i}$, and $\abstainvar_{ij}$ is $1$ if $\newresultvar_{ij} \neq \defaultlabel$.
To ensure that the accuracy for each datapoint $\x_i$ is above $\iat$, we have to ensure that out of rules that do not return \defaultlabel \xspace on $\x_i$, i.e., all $j \in [1,m]$ where $\abstainvar_{ij}=1$, at least a fraction of $\iat$ have the correct label ($\correctvar_{ij}$=1). This can be enforced if $\sum_{j} \correctvar_{ij} - \sum_j \abstainvar_{ij} \cdot \iat \leq 0$ or equivalently $\sum_j \correctvar_{ij} \geq \sum_j \abstainvar_{ij} \cdot \iat$. A symmetric condition is used to ensure \gls{lf} accuracy using the threshold $\lat$ and summing up over all datapoints instead of over all rules. Finally, we need to ensure that each datapoint $\x_i$ receives a sufficient number of labels $\neq \defaultlabel$. Recall that $\abstainvar_{ij}$ encodes whether \gls{lf} $\rl_j'$ returns a non-abstain label. Thus, for $m$ rules we have to enforce: $\forall i \in [1,n]: \sum_j \abstainvar_{ij} \geq m \cdot \inat$.
The full MILP is shown below. 
The non-linear constraints for indicator variables can be translated into linear constraints
using the so-called Big M technique~\cite{griva-08-lnoped}.


\begin{center}
\hspace{-20mm}\textbf{minimize} $\sum_{i} \sum_{j} \changedvar_{ij}$ \hspace{2mm} \textbf{subject to}\\
  \begin{minipage}{0.3\linewidth}
    \begin{align*}
      &\forall i \in [1,n], j \in [1,m]:\\
      &\hspace{5mm}\newresultvar_{ij} \in [0,\card{\outdb} - 1] \\
      &\hspace{3.5mm}\changedvar_{ij} = \indicator[\newresultvar_{ij} \neq \rl_j(\x_i)]\\
      &\hspace{5mm}\correctvar_{ij} = \indicator[\newresultvar_{ij} = \gtl{\x_i}]\\
      &\hspace{5mm}\abstainvar_{ij} = \indicator[\newresultvar_{ij} > 0]
    \end{align*}
  \end{minipage}
  \begin{minipage}{0.6\linewidth}
    \centering
    \begin{align*}
    \forall i \in [1,n]: &\sum_j \correctvar_{ij} \geq \sum_j \abstainvar_{ij} \cdot \iat\\
    \forall i \in [1,n]: &\sum_j \abstainvar_{ij} \geq m \cdot \inat\\
    \forall j \in [1,m]: &\sum_i \correctvar_{ij} \geq \sum_i \abstainvar_{ij} \cdot \lat
    \end{align*}
  \end{minipage}
\end{center}

As we show next, the solution of the MILP is a solution for the rule repair problem as long as the expected changes to the \gls{lf} results on \xcomplaints encoded in the variables $o_{ij}$ can be implemented as a repair sequence \rseq. 
As we will show in \Cref{sec:single-rule-repairs} such a repair sequence is guaranteed to exist as long as we choose the space of predicates to use in refinements carefully.

\begin{proposition}\label{prop:correctness-of-the-m}
Consider rules $\rules$, $\xcomplaints$, and the output $\newresultvar_{ij}$ produced as a solution to the \gls{milp}. If there exists a repair sequence $\rseq$ such that for $\repairrules = \rseq(\rules)$ the output on $\xcomplaints$ is equal to $\newresultvar_{ij}$ for all $i \in [1,n]$ and $j \in [1,m]$, then $\rseq$ is a solution to the \gls{rrp}.
\end{proposition}

\paratitle{MILP Size}
The number of constraints and variables in the MILP is both in $\oNotation{n \cdot m}$ where $n = \card{\xcomplaints}$ and $m = \card{\rules}$.
While solving \glspl{milp} is hard in general, we demonstrate experimentally that the runtime is acceptable for $\card{\xcomplaints} \leq 200$.


\begin{example}
Consider a set of 3 datapoints $\xcomplaints = \{\x_1, \x_2, x_3\}$ with ground truth labels $\gtl{\x_1} = 2$, $\gtl{\x_2} = 1$, $\gtl{\x_3} = 2$,  and three rules $\rl_1$ to $\rl_3$  labels $\outdb = \{0,1,2\}$ where $\defaultlabel = 0$ and assume that these rules return the results on $\xcomplaints$ shown below on the left where abstain (incorrect) labels are highlighted in blue (red).
Assume that all thresholds are set to 50\%. That is, each datapoint should receive at least two labels $\neq \defaultlabel$, and the accuracy for datapoints and rules is at least 50\% (1 correct label if 2 non-abstain labels are returned and 2 correct labels for no abstain label). The minimum number of changes required to fulfill these constraints is 4. One possible solution for the MILP is shown below on the right with modified cells (with correct labels) shown with a black background.
\end{example}
\begin{center}
  \begin{minipage}{0.45 \linewidth}
    \centering
  \begin{tabular}{|c|c|c|c|}
    \cline{2-4}
    \multicolumn{1}{c|}{} & \tabhead{$\rl_1$}   & \tabhead{$\rl_2$}  & \tabhead{$\rl_3$}   \\ \hline
    \tabhead{$\x_1$}      & \textcolor{red}{1}  & \textcolor{red}{1} & 2                   \\
    \tabhead{$\x_2$}      & \textcolor{blue}{0} & 1                  & \textcolor{blue}{0} \\
    \tabhead{$\x_3$}      & \textcolor{blue}{0} & \textcolor{red}{1} & \textcolor{blue}{0} \\
    \hline
  \end{tabular}
  \end{minipage}
  \begin{minipage}{0.5 \linewidth}
    \centering
  \begin{tabular}{|c|c|c|c|}
    \cline{2-4}
    \multicolumn{1}{c|}{$\boldsymbol{o_{ij}}$} & \tabhead{$i=1$}         & \tabhead{$i=2$}         & \tabhead{$i=3$}         \\ \hline
    \tabhead{$j=1$}                            & \cellcolor{black}\textcolor{white}{2} & \textcolor{red}{1}      & 2                       \\
    \tabhead{$j=2$}                            & \textcolor{blue}{0}     & 1                       & \cellcolor{black}\textcolor{white}{1} \\
    \tabhead{$j=3$}                            & \cellcolor{black}\textcolor{white}{2} & \cellcolor{black}\textcolor{white}{2} & \textcolor{blue}{0}     \\
    \hline
  \end{tabular}
  \end{minipage}
\end{center}


Given the outputs $\newresultvar_{ij}$ of the \gls{milp}, we need to find a repair sequence $\rseq$ such that for $\repairrules = \rseq(\rules) = \{\rl_1', \ldots, \rl_m'\}$ we have $\rl_j'(\x_i) = \newresultvar_{ij}$ for all $i \in [1,n]$ and $j \in [1,m]$.  
An important observation regarding this goal is that as rules operate independently of each other, we can solve this problem one rule at a time. 


\section{Single Rule Refinement}
\label{sec:single-rule-repairs}
We now detail our approach for refining a single rule.
\subsection{Rule Repair}
\label{sec:rule-repair}

We now formalize the problem of generating a sequence of refinement steps \rseq of minimal size for a rule $\rl_j$ such that for a set of datapoints and labels $\complaints = \{ (\x_i, \lab_i) \}_{i=1}^n$, $\rl_j' = \rseq(\rl_j)$ we have $\rl_j'(\x_i) = \lab_i$ for all $(\x_i,\lab_i) \in \complaints$. We can use this algorithm to implement the changes to rule outputs computed by the \gls{milp} from the previous section using:
\(
  \complaints = \{ (\x_i, \newresultvar_{ij}) \mid \x_i \in \xcomplaints \}.
\)
For a sequence of refinement repairs \rseq we define its cost as
%
\(
  \pcost(\rseq) = \card{\rseq}.
\)

\begin{definition}[The Single Rule Refinement 
Problem]\label{def:single-repair-problem}
Given a rule \rl, a set of datapoints with desired labels $\complaints$,
and a set of allowable predicates $\preds$, find a sequence of refinements  $\rseq_{min}$ using predicates from $\preds$ such that for $\rl_{fix} = \rseq(\rl)$ we have:
\begin{align*}
\rseq_{min} &= \argmin_{\rseq}\, \pcost(\rseq) \hspace{2mm}
 \text{\bf subject to}
& \forall (\x,\lab) \in \complaints: \rl_{fix}(\x) = \lab
\end{align*}
\end{definition}

Let $\fixp$ denote the set of paths (from the root to a label on a leaf) in rule $\rl$ that are taken by the datapoints from \cdp.
For $\apath \in \fixp$,  $\pdps{\apath}$  denotes all datapoints from $\cdp$ for which the path is $\apath$, hence also $\cdp = \bigcup_{\apath \in \fixp} \pdps{\apath}$.
Similarly, $\pathc{\apath}$ denotes the subset of $\complaints$ for datapoints \x with path $\apath$ in rule \rl.
The algorithm for solving the \gls{srrp} we will present in the following exploits two important
properties of this problem.

\noindent\textbf{Independence of path repairs.}
As any refinement in a minimal repair will only extend paths in $\fixp$ (any other refinement does not affect the labels for \cdp) and refinements at any path $\apath_1$ do not affect the labels of datapoints in $\pdps{\apath_2}$ for a path $\apath_2 \neq \apath_1$, a solution to the \gls{srrp} can be constructed one path at a time (see \iftechreport{\Cref{sec:proof-path-indendence}}\ifnottechreport{\cite{li2025refininglabelingfunctionslimited}} for the formal proof).

\noindent\textbf{Existence of path repairs.} In \iftechreport{\cref{sec:proof-crefl-repa}}\ifnottechreport{\cite{li2025refininglabelingfunctionslimited}}, we show that path repairs with a cost of at most $\card{\pdps{\apath}}$ are guaranteed to exist as long as the space of predicates is partitioning. That is, for any two datapoints $\x_1$ and $\x_2$ we can find a predicate $\pred$ such that $\pred(\x_1) \neq \pred(\x_2)$. Note that for textual data, even a simple predicate space that only contains predicates of the form $w \in \x$ where $w$ is a word is partitioning as long as any two datapoints (documents in the case of text data) will differ in at least one word. Intuitively, this guarantees the existence of a repair as for any two datapoints $\x_1$ and $\x_2$ with $\complaints(\x_1) \neq \complaints(\x_2)$ that share the same path (and, thus, also label) in a rule \rl we can refine \rl using an appropriate predicate \pred with $\pred(\x_1) \neq \pred(\x_2)$ to assign the desired labels to $\x_1$ and $\x_2$.

The pseudocode for \frepairrule is given in \Cref{algo:single-rule-repair}.
Given a single rule $\rl$, this algorithm determines a refinement-based repair $\rseq_{min}$ for \rl such that $\rseq_{min}(\rl)$ returns the designed label $\complaints(\x)$ for \emph{all} datapoints specified in \complaints by refining one path at a time using a function \fpathrefine. The problem solved by \fpathrefine is \nphard. Next, we introduce an algorithm implementing \fpathrefine that utilizes an information-theoretic heuristic that does not guarantee that the returned repair is minimal but works well in practice.

\begin{algorithm}[t]
  \SetKwInOut{Input}{Input}\SetKwInOut{Output}{Output}
  \LinesNumbered
  \Input{Rule \rl, Labelled datapoints $\complaints$. }
  \Output{Repair sequence $\rseq$ such that $\rseq(\rl)$ fixes $\complaints$}
  \BlankLine
  $\plabels \gets \emptyset, \rseq \gets \emptyset$ \\
  $\fixp \gets \{\rpath{\rl}{\x} \mid \x \in \cdp \}$\\
  $\rl_{cur} \gets \rl$\\
  \ForEach(\tcc*[h]{Fix one path at a time}){$\apath \in \fixp$}
  {\label{l:iter-path}
    \tcc{Fix path $P$ to return correctly labels on $\complaints$}
    $\pz{\apath} \gets \{ (\x,\lab) \mid  (\x,\lab) \in \complaints \land  \rpath{\rl}{\x} = \apath\}$  \label{l:rr-prepare-end}\\
    $\rop \gets \fpathrefine(\rl_{cur}, \apath, \pz{\apath})$\\
    $\rl_{cur} \gets \rop(\rl_{cur})$\\
    $\rseq \gets \rseq.\fappend(\rop)$\\
  }
\Return $\rseq$
\caption{SingleRuleRefine}
\label{algo:single-rule-repair}
\end{algorithm}


\subsection{Path Repair: \abbrEntropyPathRepair}
\label{sec:rule-refin-repa}

\begin{algorithm}[t]
  \SetKwInOut{Input}{Input}\SetKwInOut{Output}{Output}
  \LinesNumbered
  \Input{Rule \rl, Path $\apath_{in}$, Ground truth labels $\pathc{\apath_{in}}$\\
  }
  \Output{Repair sequence $\rseq$ which fixes $\rl$ wrt. $\pathc{\apath_{in}}$}
  \BlankLine
  $todo \gets [(\apath_{in},\pathc{\apath_{in}})]$\\
  $\rseq \gets []$\\
  $\rl_{cur} \gets \rl$\\
  $\preds_{all} \gets \fgetallpreds(\apath_{in},\pathc{\apath_{in}})$\\
  \While{$todo \neq \emptyset$}
  {\label{l:entropy-todo-iteration}
    $(\apath,\pathc{\apath}) \gets pop(todo)$\\
    $\pred_{new} \gets \argmin_{\pred \in \preds_{all}} \gi(\pathc{\apath},\pred)$\\
    $\asslabel_{\lfalse} \gets \{ (\x,\lab) \mid (\x, \lab) \in \pathc{\apath} \land \neg \pred(\x) \}$\\
    $\asslabel_{\ltrue} \gets \{ (\x,\lab) \mid (\x, \lab) \in \pathc{\apath}\land  \pred(\x) \}$\\
    $\lab_{max} \gets \argmax_{\lab \in \outdb} \card{\{ \x \mid \asslabel_{\ltrue}(\x) = \lab \} }$\\
    $\rop_{new} \gets \rrefine{\rl_{cur}}{\apath}{\pred}{\labOf{\apath}}{\lab_{max}}$\\
    $\rl_{cur} \gets \rop_{new}(\rl_{cur})$\\
    $\rseq \gets \rseq.append(\rop_{new})$\\
    \If{$\card{\outdb_{\asslabel_{\lfalse}}} > 1$}
    {
      $todo.push((\rpath{\rl_{cur}}{\asslabel_{\lfalse}}, \asslabel_{\lfalse}))$\\
    }
    \If{$\card{\outdb_{\asslabel_{\ltrue}}} > 1$}
    {
      $todo.push((\rpath{\rl_{cur}}{\asslabel_{\ltrue}}, \asslabel_{\ltrue}))$\\
    }
  }
  \Return $\rseq$
  \caption{\abbrEntropyPathRepair}
  \label{algo:entropy-path-repair}
\end{algorithm}

Given a rule \rl, a path $\apath_{in} \in \fixp$, and the datapoints and desired labels for this path $\pz{\apath_{in}}$,
our algorithm \emph{\abbrEntropyPathRepair} avoids the exponential runtime of an optimal brute force algorithm \abbrOptimalPathRepair that enumerates all possible refinements (see \iftechreport{\Cref{sec:abbr-opt-path-repair-algo}}\ifnottechreport{\cite{li2025refininglabelingfunctionslimited}}).
 We achieve this by greedily selecting predicates 
that best separate datapoints with different labels at each step.
To measure the quality of a split, we employ the entropy-based \emph{Gini impurity score} $\gi$~\cite{kotsiantis-13-dt}. Given a candidate predicate \pred for splitting a set of datapoints and their labels at path \apath
(\pathc{\apath}), we denote the subsets of \pathc{\apath} generated by splitting \pathc{\apath} based on \pred:
%
\begin{align*}
  \asslabel_{\lfalse} &= \{ (\x,\lab) \mid (\x, \lab) \in \pathc{\apath} \land \neg \pred(\x) \}\\
  \asslabel_{\ltrue} &= \{ (\x,\lab) \mid (\x, \lab) \in \pathc{\apath} \land  \pred(\x) \}
\end{align*}
%
Using $\asslabel_{\lfalse}$ and $\asslabel_{\ltrue}$ we define the score $\gi(\pathc{\apath},\pred)$ for  $\pred$: 
%
\begin{align*}
  \gi(\pathc{\apath},\pred) &= \frac{\card{\asslabel_{\lfalse}} \cdot \gini(\asslabel_{\lfalse}) + \card{\asslabel_{\ltrue}} \cdot \gini(\asslabel_{\ltrue})}{\card{\pathc{\apath}}}
\end{align*}\\[-7mm]
\begin{align*}
\gini(\asslabel) &= 1 - \sum_{\lab \in \outdb_{\asslabel}} p(\lab)^2 &
  p(\lab) &= \frac{\card{\{ \x \mid \asslabel(\x) = \lab \}}}{\card{\asslabel}}
\end{align*}
For a set of ground truth labels $\asslabel$,
$\gini(\asslabel)$ is minimal if $\outdb_{\asslabel} = \{ \lab \mid \exists \x: (\x,\lab) \in \asslabel \}$ contains a single label.
Intuitively, we want to select predicates such that all datapoints that reach a particular leaf node are assigned the same label. At each step, the best separation is achieved by selecting a predicate \pred that minimizes $\gi(\pathc{\apath},\pred)$.

\Cref{algo:entropy-path-repair} first determines all candidate predicates using function \fgetallpreds. Then, it iteratively selects predicates until all datapoints are assigned the expected label by the rule. For that, we maintain a queue of paths paired with a set  $\pathc{\apath}$ of datapoints with expected labels that still need to be processed. In each iteration of the algorithm's main loop, we pop one pair of a path $\apath$ and datapoints with labels $\pathc{\apath}$ from the queue. We then determine the predicate \pred that minimizes the entropy of $\pathc{\apath}$. Afterward, we determine two subsets of datapoints from $\pathdps{\apath}$:  datapoints fulfilling $\pred$ and those that do not. We then generate a refinement repair step $\rop_{new}$ for the current version of the rule ($\rl_{cur}$) that replaces the last element on $\apath_{cur}$ with predicate $\pred$ ($Y[\apath]$ denotes the label of the node at the end of $\apath$). The child at the \ltrue edge of the node for $\pred$ is then assigned the most prevalent label $\lab_{max}$  for the datapoints at this node (the datapoints from $\asslabel_{\ltrue}$). Finally, unless they only contain one label, new entries for $\asslabel_{\lfalse}$ and $\asslabel_{\ltrue}$ are appended to the queue.
%
%
As shown below, \abbrEntropyPathRepair is correct (the proof is shown in \iftechreport{\Cref{sec:proof-path-algo-correctness}}\ifnottechreport{\cite{li2025refininglabelingfunctionslimited}}).

\begin{theorem}[Correctness]\label{theo:path-repair-correct}
  Consider a rule \rl, ground-truth labels of a set of datapoints  \pathc{\apath}, and partitioning space of predicates $\preds$. Let $\rseq$ be the repair sequence produced by 
  \abbrEntropyPathRepair for path \apath.
  Then we have:\\[-6mm]
  \[
    \forall (\x,\lab) \in \pathc{\apath}: \rresult{\rseq(\rl)}{\x} = \lab
  \]
\end{theorem}
\cut{
\begin{proof}{Sketch}
The full proof is shown in \Cref{sec:proof-path-algo-correctness}.
\end{proof}
}


\section{Experiments}\label{sec:experiments}
\newcommand{\abbrLF}{LF\xspace}
\newcommand{\abbrDC}{DC\xspace}

\newcommand{\DSyoutube}{\textit{YTSpam}\xspace}
\newcommand{\DSenron}{\textit{Enron}\xspace}
\newcommand{\DSamazon}{\textit{Amazon}\xspace}
\newcommand{\DSamazonhalf}{\textit{Amazon-0.5}\xspace}
\newcommand{\DSagnews}{\textit{AGnews}\xspace}
\newcommand{\DSpp}{\textit{PP}\xspace}
\newcommand{\DSimdb}{\textit{IMDB}\xspace}
\newcommand{\DSfnews}{\textit{FNews}\xspace}
\newcommand{\DSyelp}{\textit{Yelp}\xspace}
\newcommand{\DStweets}{\textit{Tweets}\xspace}
\newcommand{\DSspam}{\textit{SMS}\xspace}
\newcommand{\DSplots}{\textit{MGenre}\xspace}
\newcommand{\DSPA}{\textit{PA}\xspace}
\newcommand{\DSPT}{\textit{PT}\xspace}
\newcommand{\DSchemprot}{\textit{CmPt}\xspace}

\newcommand{\DCFinder}{DCFinder\xspace}

\newcommand{\ALGnaive}{\textit{Greedy}\xspace}
\newcommand{\ALGentropy}{\textit{Entropy}\xspace}
\newcommand{\ALGopt}{\textit{Brute Force}\xspace}

\newcommand{\gpt}{\texttt{GPT-4o}\xspace}
\newcommand{\llama}{\texttt{Llama-3-8B-instruct}\xspace}

We evaluate the runtime of \oursys and its effectiveness in improving the accuracy of rules produced by Witan~\cite{DenhamLSN22} and \glspl{llm}. Additionally, we analyze the trade-offs introduced by the three path repair algorithms proposed in this work. Our experiments use \emph{Snorkel}~\cite{RatnerBEFWR20} as the default \abbrRBBM. To demonstrate that \oursys is agnostic to the choice of \gls{rbbm}, we also test it with alternative \gls{rbbm}s from \cite{zhang2021wrench}, measuring improvements in global accuracy. We assess both the runtime and the quality of the refinements produced by \oursys across several parameters. \oursys is implemented in Python, and all experiments are conducted on Oracle Linux Server 7.9 with 2 x AMD EPYC 7742 CPUs and 128GB RAM.

\begin{table}[htbp]
\begin{small}
    \centering
  {
  \begin{tabular}{|c|r|r|>{\footnotesize}c|c|c|}
\cellcolor{gray!40}\textbf{Dataset} & \cellcolor{gray!40}\textbf{\#row}  & \cellcolor{gray!40}\textbf{\#word} &  \cellcolor{gray!40}\textbf{$\outdb$} &  \cellcolor{gray!40}\textbf{\#$\glspl{lf}_{witan}$}  & \cellcolor{gray!40}\textbf{\#$\glspl{lf}_{llm}$} \\ \hline

\DSamazon & 200000 & 68.9 & pos/neg  & 15 & 23\\
\hline
\DSagnews & 60000 & 37.7 & busi/tech  & 9 & 21\\
\hline
\DSpp & 54476 & 55.8 & physician/prof  &  18 & 20\\
\hline
\DSimdb & 50000 & 230.7 & pos/neg   & 7 & 20\\
\hline
\DSfnews & 44898 & 405.9 & true/false & 11 & 20\\
\hline
\DSyelp & 38000 & 133.6 & neg/pos  & 8 & 20\\
\hline
\DSPT & 24588 & 62.2 & prof/teacher  & 7 & 19\\
\hline
\DSchemprot & 16075 & 27.9 & 10 relations & - & - \\ 
\hline
\DSPA & 12236 & 62.6 & painter/architect  & 10 & 18\\
\hline
\DStweets & 11541 & 18.5 & pos/neg & 16 & 18\\
\hline
\DSspam & 5572 & 15.6 & spam/ham  & 17 & 16 \\
\hline
\DSplots & 1945 & 26.5 & action/romance  & 10 & 14 \\

\hline
  \end{tabular}
}
  \caption{\abbrLF dataset statistics. }
   \vspace{-7mm}
  \label{tab:dataset_summary}
\end{small}
\end{table}

\paratitle{Datasets and rules}
The datasets used in the experiments are listed in \Cref{tab:dataset_summary}. Note that because of the complexity of and the nature of multi-class labels, the LFs we used for \DSchemprot are all from \cite{yu2020fine}, which has 26 LFs. We give a brief description of each dataset:
 \DSamazon: product reviews from Amazon and their sentiment label \cite{DBLP:conf/www/HeM16}.
 \DSagnews: categorized news articles from AG's corpus of news articles. For this dataset, we chose a binary class version from \cite{DenhamLSN22}.
 \DSpp: descriptions of biographies, each labeled as a physician or a professor \cite{DBLP:conf/fat/De-ArteagaRWCBC19}.
 \DSimdb: IMDB movie reviews \cite{DBLP:conf/acl/MaasDPHNP11}.
 \DSfnews: Fake news identification \cite{DBLP:journals/sap/AhmedTS18}.
 \DSyelp: Yelp reviews \cite{DBLP:conf/nips/ZhangZL15}.
  \DSPT: descriptions of individuals, each labeled as a professor or a teacher.\cite{DBLP:conf/fat/De-ArteagaRWCBC19}.
   \DSPA: descriptions of individuals, each labeled a painter or an architect. \cite{DBLP:conf/fat/De-ArteagaRWCBC19}.
   \DStweets: classification of tweets on disasters \cite{DBLP:conf/iscram/MouzannarRA18}.
   \DSspam: classification of SMS messages \cite{DBLP:conf/doceng/AlmeidaHY11}.
   \DSplots: movie genre classification based on plots \cite{VarmaR18}.
   \DSchemprot: chemical-protein relationship classification from \cite{krallinger2017overview}.
Unless stated otherwise, the experiments in this section are run with \abbrEntropyPathRepair. We present a detailed evaluation of these all path repair algorithms in \Cref{sec:expp-path-repair}.

\subsection{Refining labelling functions}\label{sec:exp-refining}
 In this experiment, we investigate the effects of several parameters on the performance and quality of the rules repaired with \oursys for several datasets.

\paratitle{Varying the number of labeled examples}
We evaluate how the size of \xcomplaints affects global accuracy. Global accuracy is defined as the accuracy of the labels predicted by the trained \abbrRBBM using the LFs compared to the ground truth labels. Given the limited scalability of MILP solvers in the number of variables, we used at most $\card{\xcomplaints} = 150$ datapoints.  The labeled datapoints are randomly sampled from $\indb$, with 50\% correct predictions by \abbrRBBM and 50\% wrong predictions within each sample. The reason for sampling in this manner is to provide sufficient evidence for correct predictions and predictions that need to be adjusted. Even if we have no control over the creation of $\xcomplaints$, we can achieve this by sampling from a larger set of labeled examples.
\Cref{exp_fig:global_accs_err_std} shows the global accuracy after retraining a Snorkel (\abbrRBBM) model with the rules refined by \oursys. The repairs improve the global accuracy on 8 out of 9 datasets, even for very small sample sizes. The variance of the new global accuracy also decreases as the amount of labeled examples increases.

\paratitle{Varying thresholds}
We evaluate the relationships between $\iat$, $\inat$, $\lat$ and new global accuracy. We used \DStweets with 20 labeled examples. The details of the experiments and analysis are shown in \Cref{appendix:milp-thresholds}. Based on the experiments, we recommend setting all of the thresholds to $\sim 0.7$.



\subsection{Runtime}\label{sec:exp-runtime}
Runtime breakdowns for a subset of the experiments from \Cref{sec:exp-refining} are shown in \Cref{exp_fig:runtimes}. For the breakdown of the other datasets, please refer to \Cref{appendix:runtime}. The total runtime increases as we increase the amount of labeled examples. The runtime of the refinement step is strongly correlated with the average length of the texts in the input dataset, i.e., the longer the average text length (as presented in \textit{average \# words} in \Cref{tab:dataset_summary}), the more time is required to select the best predicate using \abbrEntropyPathRepair.

It is important to note that the runtime changes for both \textit{snorkel run after refinement} and \textit{MILP} do not exhibit a strictly linear pattern. The reason for such non-linearity arises from the fact that the labeled datapoints are randomly sampled from $\indb$, and the complexity of solving the MILP problem depends on the sparsity of the solution space. The same reason applies for retraining with Snorkel using the refined rules. Some sets of labeled examples result in more complex rules even when the sample size is small, increasing the time required for Snorkel to fit a model.

\subsection{\glspl{llm} vs \oursys}\label{sec:llm_vs_rc}
In this section, we compare our approach against \glspl{llm}. We consider three setups: (i) using the \gls{llm} as a labeler (without any use of \glspl{lf}) and (ii) using the  \gls{llm} to generate \glspl{lf} based on with labeled examples; and (iii) using the \gls{llm} to repair labeling functions (refer to \Cref{append:llm-refine-detail}). For (ii), we then investigate whether \oursys can successfully improve the \glspl{lf} generated by the \gls{llm}. 

\paratitle{The \gls{llm} as a Labeler}
We compare the performance and quality of \oursys with Snorkel and \glspl{llm} as a standalone labeler. In this experiment, we use \gpt and \llama (Llama 3 8B), using a zero-shot prompt to describe the task.
Both \glspl{llm} receive the possible labels along with the sentences, but not the \glspl{lf}. 
To optimize API usage, we batch 10 datapoints per call for \gpt, whereas for \llama, we label one sentence per call to maintain response validity. Experiments were conducted on a Mac Studio (Apple M2 Max, 12-core CPU, 64GB unified memory, SSD storage). The setup with \oursys (RC for short shown in the plot) is the default setup from \Cref{sec:exp-refining}. The runtime and quality comparisons are presented in \Cref{exp_fig:3_labelers_accuracy_runtime_comparison}. We did set a 48-hour time limit, including only datasets where all three competitors completed the task within this time limit. \gpt achieves the highest accuracy in 6 out of 7 datasets. However, for dataset \DSchemprot, \oursys with Snorkel (37.9\%) outperforms \texttt{GPT4-o} (27.4\%). Upon further analysis, we speculate that GPT-4o's poor performance could stem from the specialized terminology and complex domain knowledge required for labeling in ChemProt. Unlike general-purpose datasets where \glspl{llm} excel, ChemProt contains highly domain-specific biomedical entity interactions, which may be challenging for zero-shot prompting. Using \glspl{llm} as an end-to-end labler comes at an unacceptable computational/monetary cost. The experiments with \gpt did cost \$254.42 for API usage. While we do not know the precise computational resources that were required, our local experiments with \llama, a significantly smaller model that also cannot compete with \oursys in terms of accuracy on most datasets, demonstrate the high computational cost of using an \gls{llm} for this purpose. In fact, \oursys is $\sim 2$ to $\sim 4$ orders of magnitude faster than \llama.
In summary, while large models like \gpt, but not smaller models like \llama, can achieve high accuracy as labelers, this comes at a prohibitively high computational cost. \oursys outperforms \llama in terms of accuracy on most datasets and between 56x to 1,312x in terms of runtime. Furthermore,  \glspl{lf} have the additional advantage of being inherently interpretable which is not the case for labeling with \glspl{llm}.


\begin{figure}[htbp]
  \centering
\begin{minipage}{0.6\linewidth}
    \centering
    \includegraphics[trim=0mm 3mm 0mm 5mm, width=\linewidth]{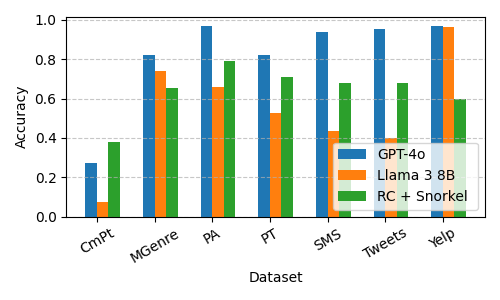}
\end{minipage}


\begin{minipage}{0.6\linewidth}
    \centering
    \includegraphics[trim=0mm 3mm 0mm 5mm, width=\linewidth]{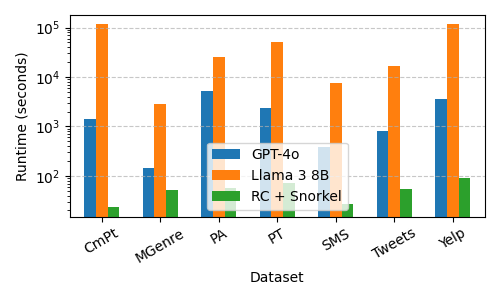}
\end{minipage}
\caption{Comparison of 3 labelers}
\vspace{-5mm}
\label{exp_fig:3_labelers_accuracy_runtime_comparison}

\end{figure}

\paratitle{The \gls{llm} as a \gls{lf} Generator}
We investigate whether \glspl{llm} can generate effective \gls{lf} and whether \oursys can fix such functions to improve their accuracy. The use of \glspl{llm} for generating \glspl{lf} has been explored in recent years \cite{DBLP:conf/edbt/GuanCK25, DBLP:conf/vldb/LiZ024}. We adapted these existing methods to generate \glspl{lf}. For an example prompt, see \iftechreport{\Cref{appendix:llm_lf_generator}}\ifnottechreport{\cite{li2025refininglabelingfunctionslimited}}. In each prompt, we sample a small set of sentences along with their ground truth labels and provide two \gls{lf} templates based on keywords and  regular expressions. Since the number of \glspl{lf} is determined by how many times we query the \gls{llm}, we scale the number of \glspl{lf} logarithmically  in  the dataset size. 
The number of \glspl{lf} produced for each dataset is reported as \#$LFs_{llm}$ in \Cref{tab:dataset_summary}.

To evaluate the quality of generated $\glspl{lf}_{witan}$, we use the same experimental setup described in \Cref{sec:exp-refining}. The results are shown in \Cref{exp_fig:global_accs_err_gpt_lf_std}. Out of the nine datasets used in this experiment, three exhibit higher original global accuracy after training with Snorkel compared to training with $\glspl{lf}_{witan}$. After refinement using labeled datapoints, we observe improvements in eight out of nine datasets.

Two datasets, \DSspam and \DStweets, show significant improvement. Upon further inspection, we observed that some of the \glspl{lf} generated by the \gls{llm} assign incorrect labels. For example, in \DSspam, one LF is defined as follows:
\pyth{return HAM if any(x in text for x in ['sorry', 'please', 'home', 'call', 'message', 'buy', 'talk', 'problem', 'help', 'ask']) else ABSTAIN}. Some of these keywords, such as ``call" and ``message", frequently appear in spam messages. \oursys successfully refines this LF by correcting its label assignment, thereby improving accuracy.

\subsection{Path Repair Algorithms}\label{sec:expp-path-repair}
Next, we compare the three path repair algorithms discussed in \Cref{sec:rule-refin-repa}. In this experiment, we used the  \DStweets dataset and randomly selected between 2 and 10 labeled examples. 
We show $\rcost$, the repaired rule size in terms of number of nodes and the runtime in \Cref{exp_fig:three_strats}. Note that for 10 labeled examples, the repair runtime for \abbrOptimalPathRepair exceeded the time limit we set for this experiment (600 secs) and, thus, is absent from the plot. The runtime of \abbrOptimalPathRepair is prohibitory large even for just 8 datapoints. \abbrEntropyPathRepair achieves almost the same repair cost as \abbrOptimalPathRepair while being significantly faster. While \abbrGreedyPredRepair is the fastest algorithm, this comes that the cost of a significantly higher repair cost.

It is obvious that the runtime for \abbrOptimalPathRepair is significantly higher than the other 2 algorithms. \abbrGreedyPredRepair is the fastest since it picks the first available predicate without any additional computation.  In terms of rule sizes after the repairs, \abbrOptimalPathRepair generates the smallest rules since it will exhaustively enumerate all the possible solutions and is guaranteed to find the smallest possible solution. It is worth noting that \abbrEntropyPathRepair is only slightly worse than \abbrOptimalPathRepair while being significantly faster.

\begin{figure*}[ht]
    \centering
    \captionsetup{aboveskip=0pt, belowskip=0pt} 
    
    \begin{subfigure}[b]{0.49\textwidth}
        \centering
        \begin{minipage}{\textwidth}
            \centering
            \begin{tabular}{@{}ccc@{}}  
                \includegraphics[trim={10 10 10 10}, clip, width=0.3\textwidth]{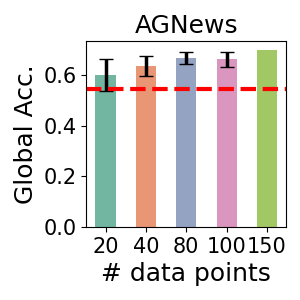} &
                \includegraphics[trim={10 10 10 10}, clip, width=0.3\textwidth]{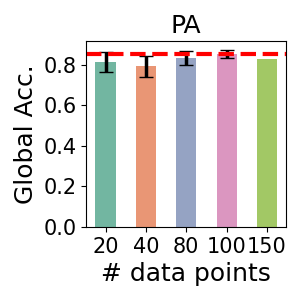} &
                \includegraphics[trim={10 10 10 10}, clip, width=0.3\textwidth]{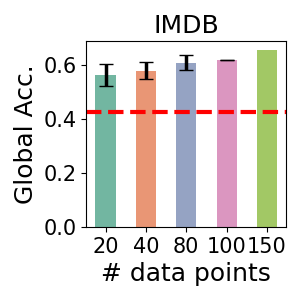} \\
                \includegraphics[trim={10 10 10 10}, clip, width=0.3\textwidth]{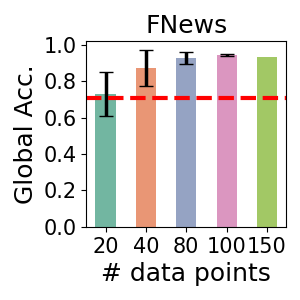} &
                \includegraphics[trim={10 10 10 10}, clip, width=0.3\textwidth]{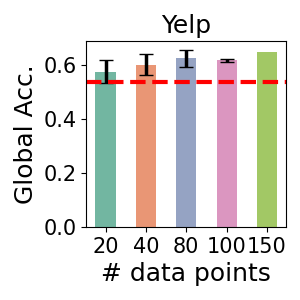} &
                \includegraphics[trim={10 10 10 10}, clip, width=0.3\textwidth]{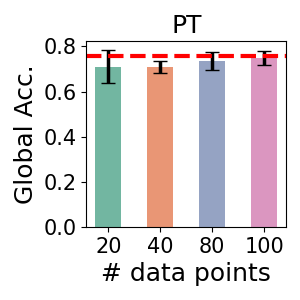} \\
                \includegraphics[trim={10 10 10 10}, clip, width=0.3\textwidth]{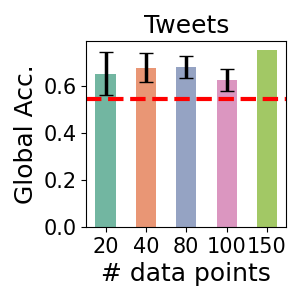} &
                \includegraphics[trim={10 10 10 10}, clip, width=0.3\textwidth]{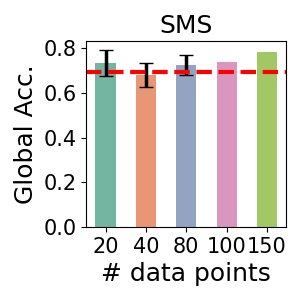} &
                \includegraphics[trim={10 10 10 10}, clip, width=0.3\textwidth]{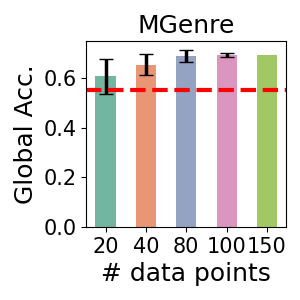} \\
            \end{tabular}
          \end{minipage}
          \vspace{-3mm}
        \caption{Change in Global accuracy, WITAN \glspl{lf}}\label{exp_fig:global_accs_err_std}        
    \end{subfigure}
    \hfill
    \begin{subfigure}[b]{0.49\textwidth}
        \centering
        \begin{minipage}{\textwidth}
            \centering
            \begin{tabular}{@{}ccc@{}}
                \includegraphics[trim={10 10 10 10}, clip, width=0.3\textwidth]{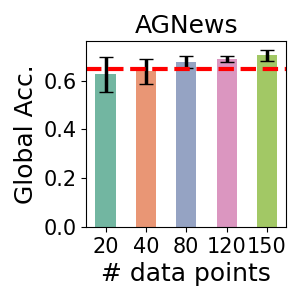} &
                \includegraphics[trim={10 10 10 10}, clip, width=0.3\textwidth]{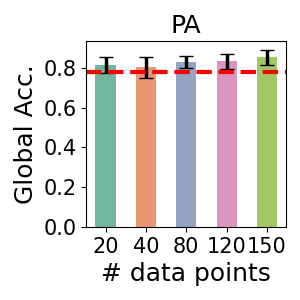} &
                \includegraphics[trim={10 10 10 10}, clip, width=0.3\textwidth]{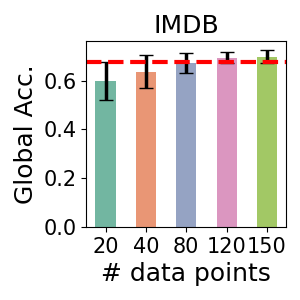} \\
                \includegraphics[trim={10 10 10 10}, clip, width=0.3\textwidth]{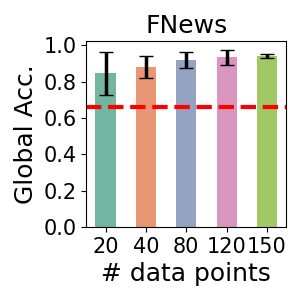} &
                \includegraphics[trim={10 10 10 10}, clip, width=0.3\textwidth]{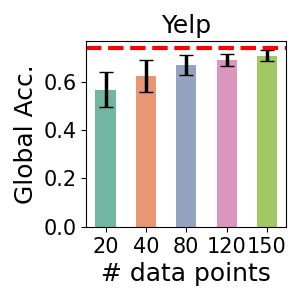} &
                \includegraphics[trim={10 10 10 10}, clip, width=0.3\textwidth]{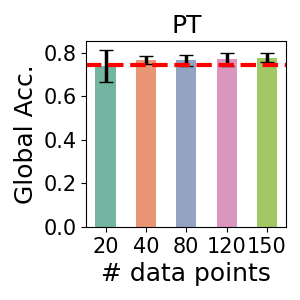} \\
                \includegraphics[trim={10 10 10 10}, clip, width=0.3\textwidth]{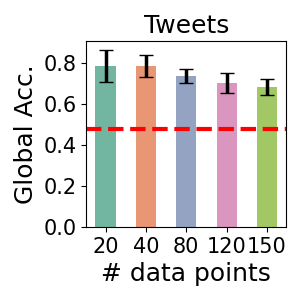} &
                \includegraphics[trim={10 10 10 10}, clip, width=0.3\textwidth]{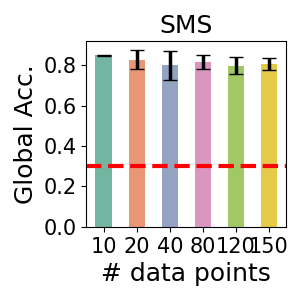} &
                \includegraphics[trim={10 10 10 10}, clip, width=0.3\textwidth]{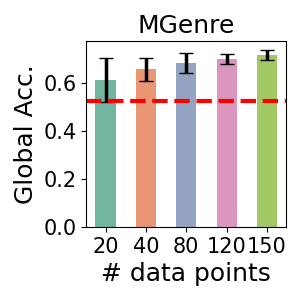} \\
            \end{tabular}
          \end{minipage}
          \vspace{-3mm}
        \caption{Change in Global accuracy, GPT-4 generated LFs}\label{exp_fig:global_accs_err_gpt_lf_std}
    \end{subfigure}
    \vspace{2mm}
    \caption{Impact of repairs on global accuracy (the red dotted line is accuracy before the repair) for \glspl{lf} generated by Witan and GPT-4. We vary the number of labeled examples \xcomplaints.}
    \vspace{-6mm}
    \label{exp_fig:global_accs}
\end{figure*}


\subsection{Complexity Evolution of Refined LFs}

To analyze the evolution of rule complexity during refinement, we conducted experiments using labeling functions generated by GPT-4o. We varied the number of labeled data points (20 and 40) and evaluated the resulting rule trees across multiple random samples. \Cref{tab:lf_complexity} reports the average tree depth and node count for three representative datasets. We observe a consistent increase in rule complexity with more input data, while the tree size (in terms of both depth and node count) grows sublinearly with respect to the input size. This suggests that the refinement process increases expressiveness efficiently without leading to overfitting.

\begin{table}[htbp]
\centering
\begin{tabular}{|c|c|c|c|}
\hline
\cellcolor{gray!40}\textbf{Dataset} & \cellcolor{gray!40}\textbf{Input Size} & \cellcolor{gray!40}\textbf{Depth} & \cellcolor{gray!40}\textbf{Node Count} \\
\hline
\DSagnews & 20 & 7.52  & 16.48 \\
\DSagnews & 40 & 11.54 & 28.06 \\
\DSimdb   & 20 & 4.95  & 11.67 \\
\DSimdb   & 40 & 7.48  & 20.21 \\
\DSspam   & 20 & 7.08  & 14.24 \\
\DSspam   & 40 & 11.08 & 25.47 \\
\hline
\end{tabular}
\vspace{2mm}
\caption{Average depth and node count of refined LFs.}
\vspace{-2mm}
\label{tab:lf_complexity}
\end{table}

\subsection{Other \gls{rbbm}s}
In addition to using Snorkel as \gls{rbbm}, we also tested models from \cite{zhang2021wrench} using the datasets from \Cref{tab:dataset_summary}. The results are shown in \Cref{tab:label_model_accuracy}. \oursys  consistently improves global accuracy across all \gls{rbbm}s, The most substantial relative gains are observed for MetaL (+15.5\%), DawidSkene (+10.5\%), and FlyingSquid (+6.9\%), while Majority Voting sees a modest improvement of +0.7\%.

\begin{table}[htbp]
\centering
\begin{tabular}{|c|c|c|c|}
\hline
\cellcolor{gray!40}\textbf{Model} & \cellcolor{gray!40}\textbf{Before} & \cellcolor{gray!40}\textbf{After} & \cellcolor{gray!40}\textbf{Rel. Gain} \\
\hline
MetaL         & 0.538 & 0.693 & +15.5\% \\
DawidSkene    & 0.566 & 0.672 & +10.5\% \\
FlyingSquid   & 0.619 & 0.688 & +6.9\% \\
Majority      & 0.685 & 0.690 & +0.7\% \\
\hline
\end{tabular}
\vspace{2mm}
\caption{Global accuracy improvements of \gls{rbbm}s after refinement.}
\label{tab:label_model_accuracy}
\vspace{-8mm}

\end{table}

\begin{figure}[t]
  \centering
    \begin{minipage}{\linewidth}
        \centering
        \includegraphics[trim={0 60 0 90}, clip, scale=0.5]{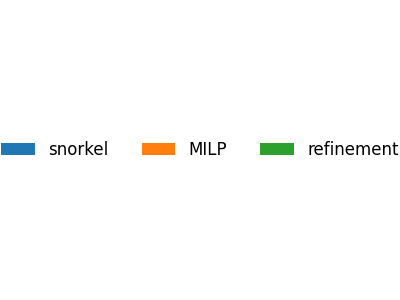}
        \vspace{-6mm} 
    \end{minipage}

    \begin{minipage}{0.45\linewidth}
        \centering
        \includegraphics[trim={5 5 5 5}, clip, scale=0.4]{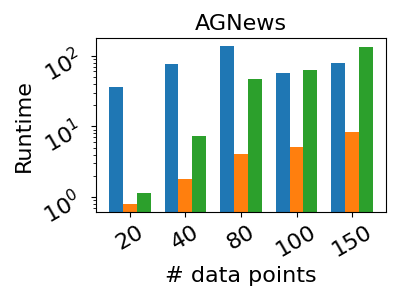}
    \end{minipage}
    \begin{minipage}{0.45\linewidth}
        \centering
        \includegraphics[trim={5 5 5 5}, clip, scale=0.4]{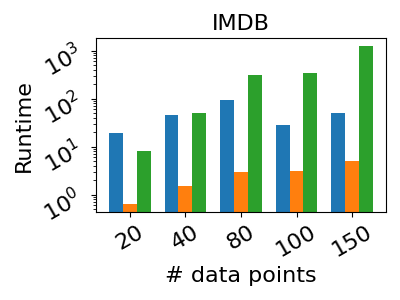}
    \end{minipage}

    \begin{minipage}{0.45\linewidth}
        \centering
        \includegraphics[trim={5 5 5 5}, clip, scale=0.4]{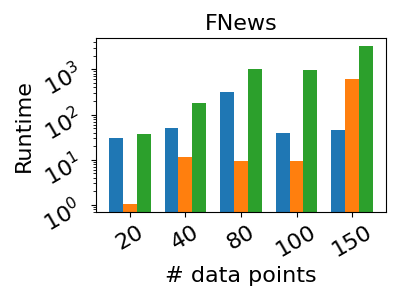}
    \end{minipage}
    \begin{minipage}{0.45\linewidth}
        \centering
        \includegraphics[trim={5 5 5 5}, clip, scale=0.4]{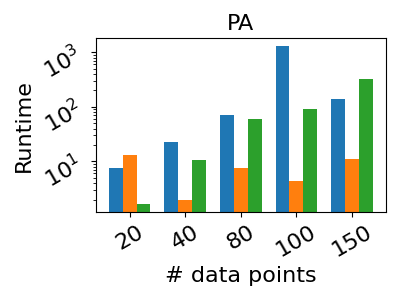}
    \end{minipage}

    \begin{minipage}{0.45\linewidth}
        \centering
        \includegraphics[trim={5 5 5 5}, clip, scale=0.4]{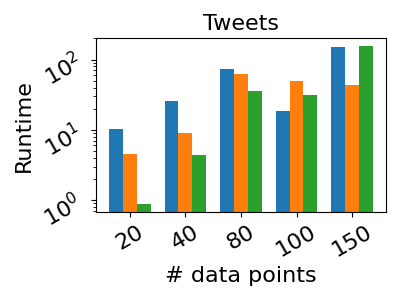}
    \end{minipage}
    \begin{minipage}{0.45\linewidth}
        \centering
        \includegraphics[trim={5 5 5 5}, clip, scale=0.4]{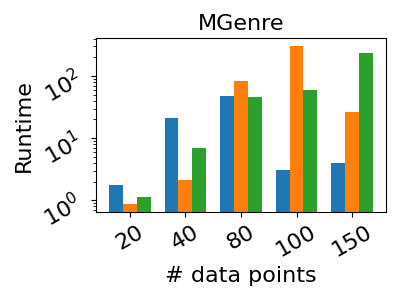}
    \end{minipage}

    \captionsetup{aboveskip=0pt, belowskip=5pt} 
    \caption{Runtime, varying the size of \xcomplaints.}
    \vspace{-5mm}
    \label{exp_fig:runtimes}
\end{figure}

\begin{figure}[h]
\centering
\begin{minipage}{0.49\linewidth} 
    \centering
    \includegraphics[trim=0mm 0mm 0mm 0mm, width=\linewidth]{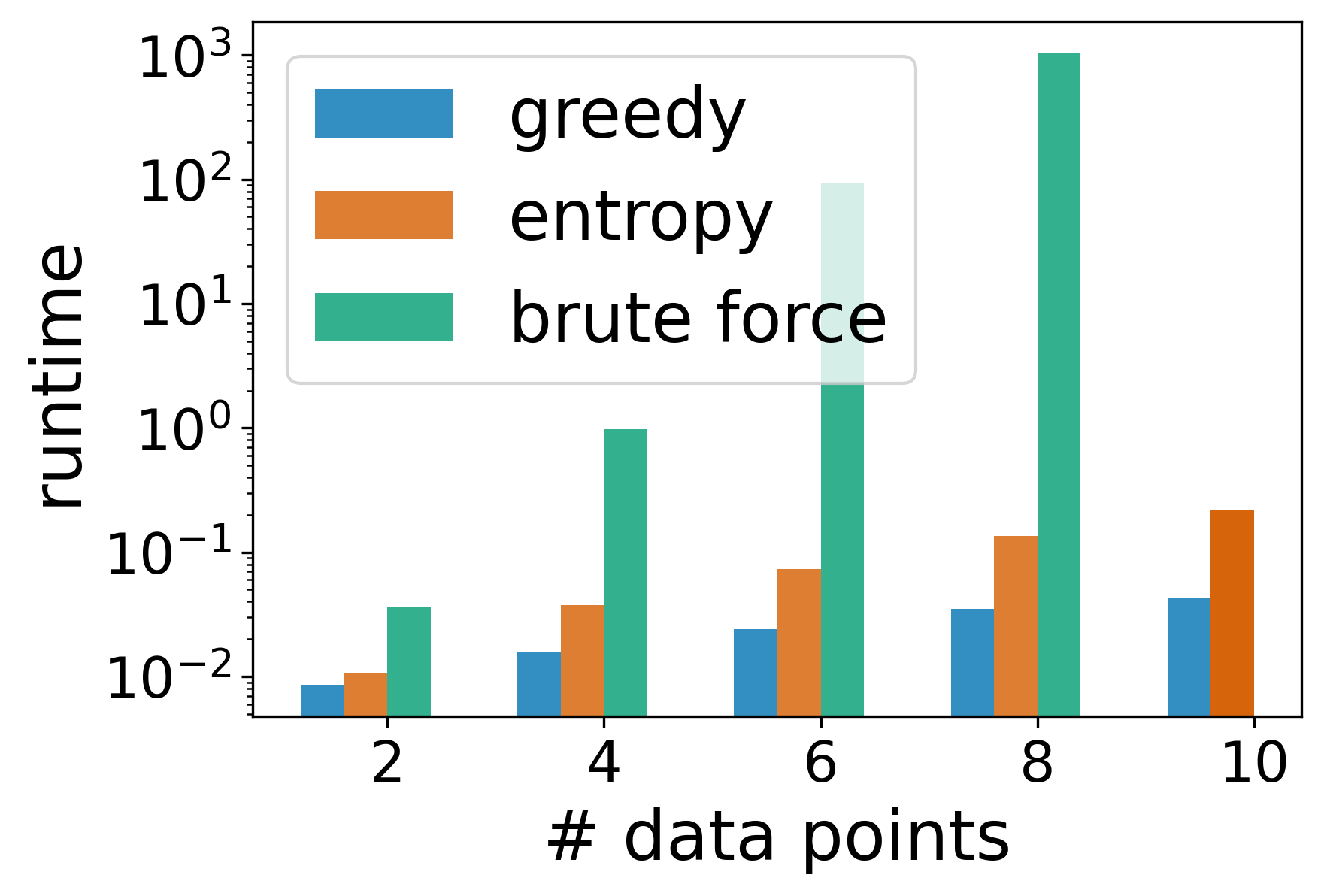}
\end{minipage}
\begin{minipage}{0.49\linewidth} 
    \centering
    \includegraphics[trim=0mm 0mm 0mm 0mm, width=\linewidth]{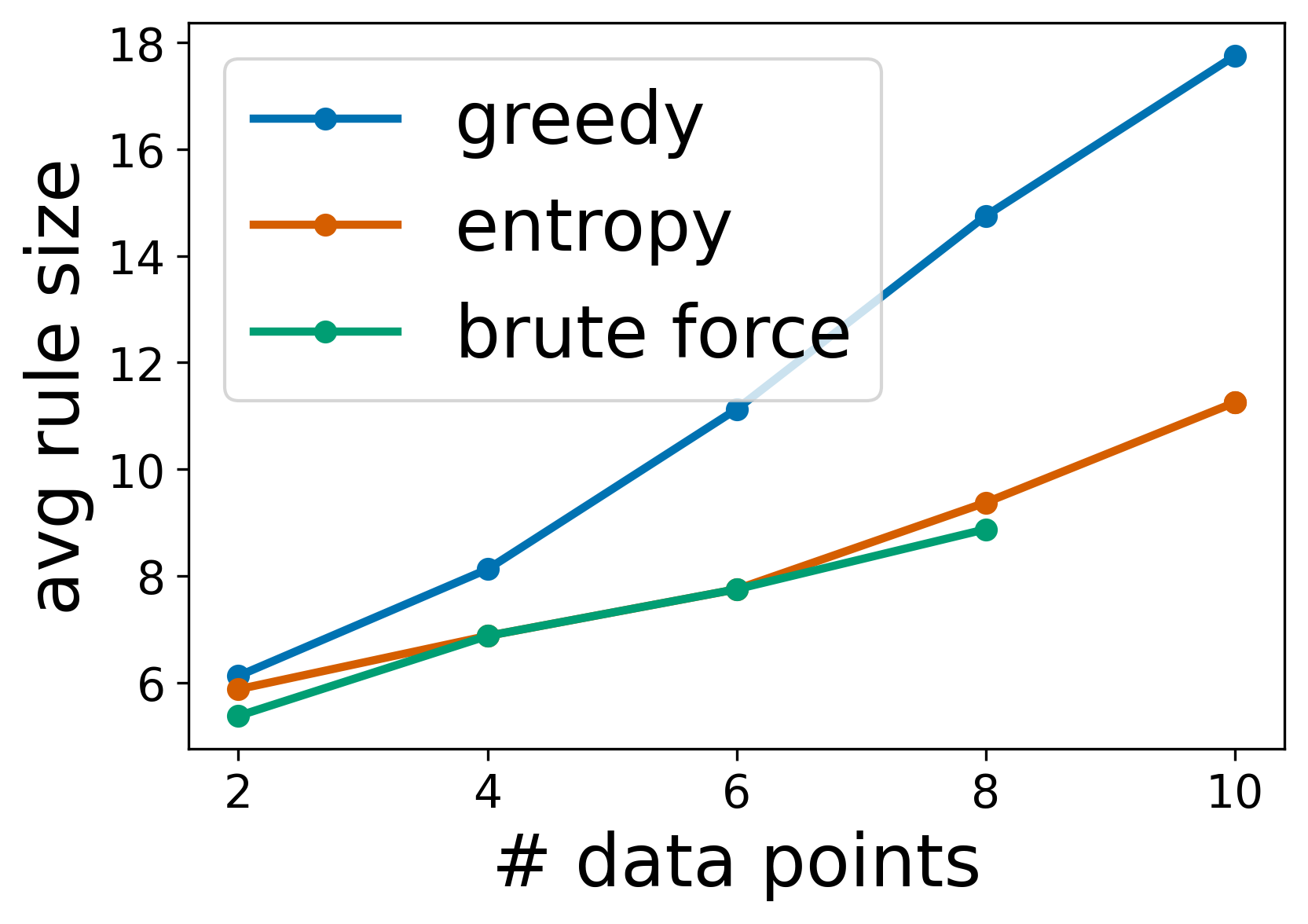}
\end{minipage}
\caption{Comparing path repair algorithms}
 \vspace{-4mm}
\label{exp_fig:three_strats}
\end{figure}


\section{Related Work}\label{sec:related}
We next survey related work on tasks that can be modeled as \abbrRBBMs 
as well as discuss approaches for automatically generating rules for \abbrRBBMs and improving a given rule set.

\paratitle{Programmatic weak supervision (\gls{pws})}
Weak supervision is a general technique of learning from noisy supervision signals, widely applied for data labeling to generate training data \cite{RatnerSWSR16,VarmaR18,RatnerBEFWR20} (the main use case we target in this work), data repair \cite{RekatsinasCIR17}, and entity matching \cite{panahi2017towards}.
Its main advantage is reducing the effort of creating training data without ground truth labels. The programmatic weak supervision paradigm pioneered in Snorkel~\cite{RatnerBEFWR20} has the additional advantage that the labeling rules are interpretable. However, as such rules are typically noisy heuristics, systems like Snorkel combine the output of \glspl{lf} using a model.
\paratitle{Automatic generation and fixing labeling functions}
While \gls{pws} proves effective, asking human annotators to create a large set of high-quality labeling functions requires domain knowledge, programming skills, and time.
As a result, the automatic generation or improvement of labeling heuristics has received much attention from the research community.
Some existing methods demand interactive user feedback in creating labeling functions~\cite{GalhotraGT21,boecking2021interactive}. \emph{Witan}~\cite{DenhamLSN22} asks a domain expert to select the automatically generated \glspl{lf} and assign labels to the \glspl{lf}. While the \glspl{lf} produced by Witan are certainly useful, we demonstrate in our experimental evaluation that applying \oursys to  Witan \glspl{lf} can significantly improve accuracy. 
Other methods generate \glspl{lf} without requiring user annotations. \emph{Snuba}~\cite{VarmaR18} fits classification models, such as decision trees and logistic regressions, as \glspl{lf} on a small labeled training set, followed by a pruning process to determine which \glspl{lf} for final use.
\emph{Datasculpt}~\cite{DBLP:conf/edbt/GuanCK25} prompts a large language model (LLM) with a small set of labeled training data and keyword- or pattern-based \glspl{lf} as in-context examples. The LLM then generates \glspl{lf} for unlabeled examples based on this input. Evaporate~\cite{arora2023language} uses an LLM to generate data extraction functions, and then it applies weak supervision to filter out low-quality functions and aggregate the results.


\cut{
Although \emph{Datasculpt} filters the generated \glspl{lf} based on accuracy and diversity, quality issues remain, and \oursys{} can further improve the LF accuracy, as discussed in Section~\ref{exp:llm}.}

Hsieh et al.~\cite{hsieh-22-n-a} propose \emph{Nemo}, a framework for selecting data to guide users in developing \glspl{lf}. It estimates the likelihood of users proposing specific \glspl{lf} using a utility metric for \glspl{lf} and a model of user behavior. 
Nemo tailors \glspl{lf} to the neighborhood of the data,
assuming that user-developed \glspl{lf} are more accurate for data similar to those used for \glspl{lf} creation.
However, unlike \oursys, Nemo lacks a mechanism for the user to provide feedback on the labeling results, preventing the automatic deletion and refinement of \glspl{lf}.
\emph{ULF}~\cite{abs-2204-06863} is an unsupervised system for adjusting \glspl{lf} assignment for unlabeled samples (instead of repairing them) using k-fold cross-validation, 
extending previous approaches addressing labeling errors~\cite{WangSLLLH19,NorthcuttJC21}.

\paratitle{Explanations for weakly supervised systems}
There is a large body of work on explaining the results of weak-supervised systems that target improving the final model or better involving human annotators~\cite{zhang-22-unpwssinf,zhang2023labelvizier,yu2023alfred,butcher2023optimising,boecking2021interactive,zhang-22-unpwssinf,guan-24-w}. For instance, \cite{zhang-22-unpwssinf} uses influence function to identify \glspl{lf} responsible for erroneous labels; WeShap~\cite{guan-24-w} measures the shapley value of \glspl{lf} to rank and prune \glspl{lf}. 
However, most of this work has stopped short of repairing the rules in a \abbrRBBM and, thus, are orthogonal to our work. 
Still, explanations provided by such systems might guide users in selecting what datapoints to label. 
People have also studied using human-annotated natural language explanations to build \glspl{lf}~\cite{hancock2018training}.

\section{Conclusions and Future Work}\label{sec:concl-future-work}
We study repairs for \glspl{lf} in \gls{pws} based on a small set of labeled examples. 
Our algorithm is highly effective in improving the accuracy of \abbrRBBMs by improving rules created by a human expert or automatically discovered by
a system like Witan~\cite{DenhamLSN22}.
In future work, we will explore the application of our rule repair algorithms to other tasks that can be modeled as \abbrRBBM, e.g., information extraction based on user-provided rules~\cite{RC13, LC10}. 


\section{Acknowledgments}
This work is supported in part by NSF Awards IIS-2420577, IIS-2420691, and IIS-2147061, and by the Natural Sciences and Engineering Research Council of Canada (NSERC) under award numbers RGPIN-2025-04724 and DGECR-2025-00373. The work of Amir Gilad was funded by the Israel Science Foundation (ISF) under grant 1702/24, the Scharf-Ullman Endowment, and the Alon Scholarship.

\bibliographystyle{ACM-Reference-Format}
\bibliography{bibtex}
\balance

\appendix




\iftechreport{
\section{Translating Labeling Functions Into Rules}\label{sec:rule-conversion}

In this section, we detail simple procedures for converting 
labeling functions to our rule representation (\Cref{def:rule}).

\begin{definition}[Rule]\label{def:rule}
 A rule $\rl$ over atomic predicates $\preds$ is a labeled directed binary tree where the internal nodes are predicates in $\preds$, leaves are labels from $\outdb$, and edges are marked with \ltrue and \lfalse.

A rule \rl takes as input a datapoint $\x \in \indb$ and returns a label $\rresult{\rl}{\x} \in \outdb$ for this assignment.

Let $\rroot{\rl}$ denote the root of the tree for rule \rl and let $\tchild{\rnode}$ ($\fchild{\rnode}$) denote the child of node $\rnode$ adjacent to the outgoing edge of $\rnode$ labeled \ltrue (\lfalse).
Given a datapoint $\x$, the result of rule \rl for \x is $\rresult{\rl}{\x} = \rooteval{\rl}{\x}$. Function $\reval{\cdot}{\cdot}$ operates on nodes $\rnode$ in the rule's tree and is recursively defined as follows:
\[
    \areval =
  \begin{cases}
      \lab                      & \text{if } \rnode \text{ is a leaf labeled } \lab \in \outdb \\
    \reval{\tchild{\rnode}}{\x} & \text{if } \rnode(\x) \text{ is true}
                                                                                               \\
    \reval{\fchild{\rnode}}{\x} & \text{if } \rnode(\x) \text{ is false}
                                                                                               \\
  \end{cases}
\]
Here $\rnode(\x)$ denotes replacing variable \var in the predicate of node $\rnode$ with \x and evaluating the resulting predicate.
\end{definition}




\IncMargin{1em}
\begin{algorithm}[t]
\caption{Convert LF to Rule}
\label{algo:lf-to-rule}
    \SetKwInOut{Input}{Input}\SetKwInOut{Output}{Output}
    \LinesNumbered
    \Input{The code $C$ of a LF $f$}
    \Output{$r_f$, a rule representation of $f$} \BlankLine
    $V,E = \emptyset, \emptyset$\;
    $r_f \gets (V,E)$\;
    Let $C$ be the source code of $f$\;
    \SetKwProg{proc}{}{}{}
    $\text{\textbf{LF-to-Rule}}(r, C)$:
    \proc{}{
      \If{$C = $\texttt{if cond: B1 else: B2}{if cond: B1 else: B2} $\land$ \texttt{ispure}(cond) $\land$ \texttt{purereturn}(B1) $\land$ \texttt{purereturn}(B2)}
      {
        $v_{true}  \gets $ \texttt{LF-to-Rule} $(B1)$\\
        $v_{false}  \gets$ \texttt{LF-to-Rule} $(B2)$\\
        $\rnode_{root} \gets $ \texttt{Pred-To-Rule} $(cond, \node_{false}, \node_{true})$
      }
      \ElseIf{$C = \texttt{if cond: B1 else: B2}{return y} \land \lab \in \outdb$}
      {
        $\rnode_{root} = \lab$
      }
      \Else{
        $\rnode_{root} \gets \mathtt{Translate\mbox{-}BBox}(C)$
      }
      \Return $\rnode_{root}$
    }
    $\text{\textbf{Pred-to-Rule}}(\psi, \rnode_{false}, \rnode_{true})$:
    \proc{}{
      \If{$\psi = \psi_1 \lor \psi_2$}
      {
       $\rnode_{left} \gets $ \texttt{Pred-to-Rule} $(\psi_2, \rnode_{false}, \rnode_{true})$\\
       $
  \rnode_{root} \gets $    \begin{tikzpicture}[baseline=-12pt,scale=.5,
      level 1/.style={level distance=1.4cm, sibling distance=2cm},
      snode/.style = {shape=rectangle, rounded corners, draw, align=center, top color=white, bottom color=red!20},
      newsnode/.style = {shape=rectangle, rounded corners, draw, align=center, top color=white, bottom color=green!20},
      logic/.style = {shape=rectangle, rounded corners, draw, align=center, top color=toplogiccolor, bottom color=bottomlogiccolor},
      newlogic/.style = {shape=rectangle, rounded corners, draw, align=center, top color=white, bottom color=green!20},-]
      \node[logic](a) {$\psi_1$}
        child {node[logic](b) {$\node_{left}$} 
            edge from parent node[rectangle,left, near start] {{\small\lfalse}}}
        child {node[logic](c) {$\node_{true}$} 
            edge from parent node[rectangle,right, near start] {{\small\ltrue}}};
    \end{tikzpicture}
      }
      \ElseIf{$\psi = \psi_1 \land \psi_2$}
      {
       $\node_{right} \gets $ \texttt{Pred-to-Rule} $(\psi_2, \node_{false}, \node_{true})$\\
        $
  \rnode_{root} \gets $    \begin{tikzpicture}[baseline=-12pt,scale=.5,
      level 1/.style={level distance=1.5cm, sibling distance=2cm},
      snode/.style = {shape=rectangle, rounded corners, draw, align=center, top color=white, bottom color=red!20},
      newsnode/.style = {shape=rectangle, rounded corners, draw, align=center, top color=white, bottom color=green!20},
      logic/.style = {shape=rectangle, rounded corners, draw, align=center, top color=toplogiccolor, bottom color=bottomlogiccolor},
      newlogic/.style = {shape=rectangle, rounded corners, draw, align=center, top color=white, bottom color=green!20},-]
    \node[logic](a) {$\psi_1$}
        child {node[logic](b) {$\node_{false}$} 
            edge from parent node[rectangle,left, near start] {{\small\lfalse}}}
        child {node[logic](c) {$\node_{true}$} 
            edge from parent node[rectangle,right, near start] {{\small\ltrue}}};
    \end{tikzpicture}
      }
      \Else
      {
        $ \rnode_{root} \gets$     \begin{tikzpicture}[baseline=-12pt,scale=.5,
      level 1/.style={level distance=1.5cm, sibling distance=2cm},
      snode/.style = {shape=rectangle, rounded corners, draw, align=center, top color=white, bottom color=red!20},
      newsnode/.style = {shape=rectangle, rounded corners, draw, align=center, top color=white, bottom color=green!20},
      logic/.style = {shape=rectangle, rounded corners, draw, align=center, top color=toplogiccolor, bottom color=bottomlogiccolor},
      newlogic/.style = {shape=rectangle, rounded corners, draw, align=center, top color=white, bottom color=green!20},-]
      \node[logic](a) {$\psi$}
        child {node[logic](b) {$\node_{false}$} 
            edge from parent node[rectangle,left, near start] {{\small\lfalse}}}
        child {node[logic](c) {$\node_{true}$} 
            edge from parent node[rectangle,right, near start] {{\small\ltrue}}};
    \end{tikzpicture}
      }
      \Return $\rnode_{root}$
    }
    $\text{\textbf{Translate-BBox}}(C):$
    \LinesNumbered
    \proc{}{
      $\node_{root} = \defaultlabel$\;
      \For{$\lab \in \outdb / \{\defaultlabel\}$}
      {
  $\rnode_{root} \gets     \begin{tikzpicture}[baseline=-12pt,scale=.5,
      level 1/.style={level distance=1.5cm, sibling distance=2cm},
      snode/.style = {shape=rectangle, rounded corners, draw, align=center, top color=white, bottom color=red!20},
      newsnode/.style = {shape=rectangle, rounded corners, draw, align=center, top color=white, bottom color=green!20},
      logic/.style = {shape=rectangle, rounded corners, draw, align=center, top color=toplogiccolor, bottom color=bottomlogiccolor},
      newlogic/.style = {shape=rectangle, rounded corners, draw, align=center, top color=white, bottom color=green!20},-]
      \node[logic](a) {$C(\var) = \lab$}
        child {node[logic](b) {$\node_{root}$} 
            edge from parent node[rectangle,left, near start] {{\small\lfalse}}}
        child {node[logic](c) {$\lab$} 
            edge from parent node[rectangle,right, near start] {{\small\ltrue}}};
    \end{tikzpicture}$\\
      }
      \Return $\rnode_{root}$
    }
\end{algorithm}

As mentioned before, we support arbitrary labeling functions written in a general-purpose programming language. Our implementation of the translation into rules is for Python functions, as supported in Snorkel. Detecting the if-then-else rule structure and logical connectives that are supported in our rules for an arbitrary Python function is undecidable in general (can be shown through a reduction from program equivalence). However, our algorithm can still succeed in any LF written in Python if we can live if we treat every code block that our algorithm does not know how to compose as a black box that we call repeatedly on the input and compare its output against all possible labels from \outdb. In the worst case, we would wrap the whole LF in this way. Note that this does not prevent us from refining such labeling functions. However, there are several advantages in decomposing an LF into a tree with multiple predicates: (i) such a rule will make explicit the logic of the LF and, thus, may be easier to interpret by a use,r and (ii) during refinement we have more information to refine the rule as there may be multiple leaf nodes corresponding to a label \lab, each of which corresponds to a different set of predicates evaluating to true.

Our translation algorithm knows how to decompose a limited number of language features into predicates of a rule. As mentioned above, any source code block whose structure we cannot further decompose will be treated as a blackbox and will be wrapped as a predicate whose output we compare against every possible label from \outdb. Furthermore, when translating Boolean conditions, i.e., the condition of an if statement, we only decompose expressions that are logical connectives and treat all other subexpressions of the condition as atomic. While this approach may sometimes translate parts of a function's code into a black-box predicate, most \glspl{lf} we have observed in benchmarks and real applications of data programming can be decomposed by our approach. Nonetheless, our approach can easily be extended to support additional structures if need be.

\paratitle{Translating labeling functions}
Pseudo code for our algorithm is shown in \Cref{algo:lf-to-rule}. Function \texttt{LF-to-Rule} is applied the code \texttt{C} of the body of labeling function $f$. We analyze code blocks using the standard libraries for code introspection in Python, i.e., Python's AST library. If the code is an if then else condition (for brevity we do not show the the case of an if without else and other related cases) that fulfills several additional requirements, then we call \texttt{LF-to-Rule} to generate rule trees to the if and the else branch. Afterwards, we translate the condition using function \texttt{Pred-To-Rule} described next that takes as input a condition $\psi$ and the roots of subtrees to be used when the condition evaluates to false or true, respectively. For this to work, several conditions have to apply: (i) both \texttt{B1} and \texttt{B2} have to be pure, i.e., they are side-effect free, and return a label for every input. This is checked using function \texttt{purereturn}. This is necessary to ensure that we can translate \texttt{B1} and \texttt{B2} into rule fragments that return a label. Furthermore, \texttt{cond} has to be pure (checked using function \texttt{ispure}). Note that both \texttt{ispure} and \texttt{purereturn} have to check a condition that is undecidable in general. Our implementations of these functions are is sound, but not complete. That is, we may fail to realize that a code block is pure (and always returns a label in case of \texttt{purereturn}, but will never falsely claim a block to have this property.

If the code block returns a constant label \lab, then it is translated into a rule fragment with a single node $\lab$.
Finally, if the code block \texttt{C} is not a conditional statement, then we fall back to use our black box translation technique (function \texttt{Translate-BBox} explained below).

\paratitle{Translating predicates}
\texttt{Pred-to-Rule}, our function for translating predicates (Boolean conditions as used in if statements), takes as input a condition $\psi$ and the roots of two rule subtrees ($\node_{false}$ and $\node_{true}$) that should be used to determine an inputs label based on whether $\psi$ evaluates to false (true) on the input. The function checks whether the condition is of the form $\psi_1 \lor \psi_2$ or $\psi_1 \land \psi_2$. If that is the case, we decompose the condition and create an appropriate rule fragment implementing the disjunction (conjunction). For disjunctions $\psi_1 \lor \psi_2$, the result of $\node_{true}$ should be returned if $\psi_1$ evaluates to true. Otherwise, we have to check $\psi_2$ to determine whether $\node_{false}$'s or $\node_{true}$'s result should be returned. For that we generate a rule fragments as shown below where $\node_{left}$ denote the root of the rule tree generated by calling \texttt{Pred-to-Rule} to translate $\psi_2$.

\begin{center}
  \begin{tikzpicture}[baseline=-12pt,scale=.7,
      level 1/.style={level distance=1.2cm, sibling distance=1.5cm},
      snode/.style = {shape=rectangle, rounded corners, draw, align=center, top color=white, bottom color=red!20},
      newsnode/.style = {shape=rectangle, rounded corners, draw, align=center, top color=white, bottom color=green!20},
      logic/.style = {shape=rectangle, rounded corners, draw, align=center, top color=toplogiccolor, bottom color=bottomlogiccolor},
      newlogic/.style = {shape=rectangle, rounded corners, draw, align=center, top color=white, bottom color=green!20},-]
        \node[logic](a) {$\psi_1$}
        child {node[logic](b) {$\node_{left}$} 
            edge from parent node[rectangle,left, near start] {{\small\lfalse}}}
        child {node[logic](c) {$\node_{true}$} 
            edge from parent node[rectangle,right, near start] {{\small\ltrue}}};
    \end{tikzpicture}
\end{center}

The case for conjunctions is analog. If $\psi_1$ evaluates to false we have to return the result of $\node_{false}$. Otherwise, we have to check $\psi_2$ to determine whether to return $\node_{false}$'s or $\node_{true}$'s result. This is achieved using the rule fragment shown below where $\node_{right}$ denotes the root of the tree fragment generated for $\psi_2$ by calling \texttt{Pred-to-Rule} on $\psi_2$.

\begin{center}
 \begin{tikzpicture}[baseline=-12pt,scale=.7,
      level 1/.style={level distance=1.2cm, sibling distance=1.5cm},
      snode/.style = {shape=rectangle, rounded corners, draw, align=center, top color=white, bottom color=red!20},
      newsnode/.style = {shape=rectangle, rounded corners, draw, align=center, top color=white, bottom color=green!20},
      logic/.style = {shape=rectangle, rounded corners, draw, align=center, top color=toplogiccolor, bottom color=bottomlogiccolor},
      newlogic/.style = {shape=rectangle, rounded corners, draw, align=center, top color=white, bottom color=green!20},-]
      
      \node[logic](a) {$\psi_1$}
        child {node[logic](b) {$\node_{false}$} 
            edge from parent node[rectangle,left, near start] {{\small\lfalse}}}
        child {node[logic](c) {$\node_{true}$} 
            edge from parent node[rectangle,right, near start] {{\small\ltrue}}};
    \end{tikzpicture}

\end{center}

If $\psi$ is neither a conjunction nor disjunction, then we just add predicate node for the whole condition $\pred$:

\begin{center}
\begin{tikzpicture}[baseline=-12pt,scale=.7,
      level 1/.style={level distance=1.2cm, sibling distance=1.5cm},
      snode/.style = {shape=rectangle, rounded corners, draw, align=center, top color=white, bottom color=red!20},
      newsnode/.style = {shape=rectangle, rounded corners, draw, align=center, top color=white, bottom color=green!20},
      logic/.style = {shape=rectangle, rounded corners, draw, align=center, top color=toplogiccolor, bottom color=bottomlogiccolor},
      newlogic/.style = {shape=rectangle, rounded corners, draw, align=center, top color=white, bottom color=green!20},-]
      
      \node[logic](a) {$\psi$}
        child {node[logic](b) {$\node_{false}$} 
            edge from parent node[rectangle,left, near start] {{\small\lfalse}}}
        child {node[logic](c) {$\node_{true}$} 
            edge from parent node[rectangle,right, near start] {{\small\ltrue}}};
    \end{tikzpicture}
\end{center}

\paratitle{Translating blackbox code blocks}
Function \texttt{Translate-BBox} is used to translate a code block \texttt{C} that takes as input a datapoint $\x$ (assigned to variable $\var$), treating the code block as a black box. This function creates a rule subtree that compares the output of $C$ on $\var$ against every possible label $\lab \in \outdb$. Each such predicate node has a true child that is $\lab$, i.e., the rule fragment will return $\lab$ iff $C(\x) = \lab$. Note that this translation can not just be applied to full labeling functions, but also code blocks within a labeling function's code that our algorithm does not know how to decompose into predicates. \Cref{fig:translate-blackbox} shows the structure of the generated rule tree produced by \texttt{Translate-BBox} for a set of labels $\outdb = \{\lab_1, \ldots, \lab_k, \abstain \}$ where $\abstain$ is the default label ($\defaultlabel$).

\begin{figure}[thb]
    \centering
    \begin{tikzpicture}[scale=.8,
    level distance=0.2cm,  
    sibling distance=2.5cm,]
        
    \node[logic] (root) {$f(\var) = \lab_1$}
    child {node[draw=none] (dots) {$\ldots$}
        child {node[logic] (n2) {$f(\var) = \lab_{k-1}$} 
            child {node[logic] (n3) {$f(\var) = \lab_k$} 
                child {node[snode] (abstain) {$\abstain$} 
                    edge from parent node[rectangle, above left=1pt] {{\small\lfalse}}}
                child {node[snode] (labk) {$\lab_k$} 
                    edge from parent node[rectangle, above right=1pt] {{\small\ltrue}}}
                edge from parent node[rectangle, above] {{\small\lfalse}} 
            }
            child {node[snode] (labkm1) {$\lab_{k-1}$} 
                edge from parent node[rectangle, above] {{\small\ltrue}}}
            edge from parent node[rectangle, above] {{\small\lfalse}} 
        }
        child {node[snode] (labkm2) {$\lab_{k-2}$} 
            edge from parent node[rectangle, above] {{\small\ltrue}}}
        }   
        child {node[logic] (n1) {$\lab_1$} 
        edge from parent node[rectangle, above] {{\small\ltrue}}}
        ;
    \end{tikzpicture}
    \caption{Translating a blackbox function} \label{fig:translate-blackbox}
\end{figure}

As mentioned above, this translation process produces a valid rule tree $r_f$ that is equivalent to the input \gls{lf} $f$ in the sense that it returns the same results as $f$ for every possible input. Furthermore, the translation runs in \ptime. In fact, it is linear in the size of the input.

\begin{proposition}\label{prop:rule-translation-cor}
  Let $f$ be a python labeling function and let $r_f = \texttt{LF-to-Rule}(f)$. Then for all datapoints $\x$ we have
  \[
    f(\x) = r_f(\x)
  \]
The function \texttt{LF-to-Rule}'s runtime is linear in the size of $f$.
\end{proposition}
\begin{proof}[Proof Sketch]
The result is proven through induction over the structure of a labeling function.
\end{proof}
\begin{example}[Translation of Complex Labeling Functions]\label{ex:translation-of-complex-la}
  Consider the labeling function implemented in Python shown below. This function assigns label \positive to each datapoint (sentences in this example) containing the word star or stars. For sentences that do not contain any of these words, the function uses a function \pyth{sentiment_analysis} to determine the sentence's sentiment and return \positive if it is above a threshold. Otherwise, \abstain is returned. Our translation algorithm identifies that this function implements an if-then-else condition. The condition if pure and both branches are pure and return a label for every input. Thus, we translate both branches using \texttt{LF-to-Rule} and the condition using  \texttt{Pred-to-Rule}. The if branch is translated into a rule fragment with a single node $\positive$. The else branch contains assignments that our approach currently does not further analyze and, thus, is treated as a blackbox by wrapping it in a new function, say \pyth{blackbox_lf} (shown below), whose result is compared against all possible labels. Finally, the if statement's condition is translated with \texttt{Pred-to-Rule}. We show the generated rule in \Cref{fig:translate-partial-blackbox}. Note that technically the comparison of the output of \pyth{blackbox_lf} with label \negative is unnecessary as this function does not return this label for any input. This illustrates the trade-off between adding additional complexity to the translation versus simplifying the generated rules.


\begin{python}
def complex_lf(v):
  if ['star', 'stars'].intersection(v):
     return POSITIVE
  else:
     sentiment = sentiment_analysis(v)
     return POSITIVE if sentiment > 0.7 else ABSTAIN
\end{python}

\begin{python}
def blackbox_lf(v):
    sentiment = sentiment_analysis(v)
    return POSITIVE if sentiment > 0.7 else ABSTAIN
\end{python}

\end{example}

\begin{figure}[t]
    \centering
    \begin{tikzpicture}[scale=.8,
    level 1/.style={level distance=1.5cm, sibling distance=3cm},
    level 2/.style={level distance=1.5cm,sibling distance=3cm},
    level 3/.style={level distance=1.5cm, sibling distance=3cm},
    level 4/.style={level distance=1.5cm, sibling distance=2cm},
     level 5/.style={level distance=1.5cm, sibling distance=2cm}]

    \node[logic] {\pyth{['star', 'stars'].intersection(v)}}
        child {
        node[logic]{\pyth{blackbox\_lf}(\var) = \positive}
          child {
          node[logic]{\pyth{blackbox\_lf}(\var) = \negative} 
            child {
            node[snode]{$\abstain$} edge from parent node[above] {{\small\lfalse}}
            }
            child {node[snode] {$\negative$} edge from parent node[above] {{\small\ltrue}}}
            edge from parent node[above]  {{\small\lfalse}}
          } 
          child {node[snode] {$\positive$} edge from parent node[above]  {{\small\ltrue}}} 
          edge from parent node[above]  {{\small\lfalse}}
        }
        child {node[snode] {$\positive$} edge from parent node[above]  {{\small\ltrue}}};
    \end{tikzpicture}
    \caption{Translating a LF wrapping parts into a blackbox function} \label{fig:translate-partial-blackbox}
\end{figure}


\section{Proof of Theorem \ref{theo:rule-repair-hardness}}\label{sec:proofs}

We now prove the hardness of the rule repair problem. Afterward, we demonstrate that even repairing a single rule with refinement to produce a specific result on a set of datapoints \nphard if the goal is to minimize the number of predicates that are added to the rule.

\begin{proof}[Proof of \Cref{theo:rule-repair-hardness}]\label{proof:rule set repair hardness}
  We prove the theorem through a reduction from the set cover problem. Recall that the set cover problem is: given a set $U = \{e_1, \ldots, e_n\}$ and sets $S_1$ to $S_m$ such that $S_i \subseteq U$ for each $i$, does there exist $i_1$, \ldots, $i_k$ such that $\bigcup_{j=1}^{k} S_{i_j} = U$.

  Based on an instance of the set cover problem, we construct an instance of the rule repair problem as follows:
\begin{itemize}
\item The instance $\xcomplaints$ contains the following datapoints:
  \begin{itemize}
  \item $b_i$ for $i \in [1,M]$ where $M$ is a large constant, say $M = 3 \cdot n \cdot m^2$ and $\forall i: b_i \not\in U$.
  \item $a_{ij}$ for $i \in [1,m]$ and $j \in [1,n\cdot m]$ and $\forall i,j: a_{ij} \not\in U$
  \item $e_{i}$ for $e_i \in U$
  \end{itemize}
\item Labels $\outdb = \{  out, in \}$
\item Rules $\rules = \{\rl_i \}$ for $i \in [1,m]$ where $\rl_i$ has a single predicate $p_i: v \in S_i \lor \exists j: v = a_{ij}$  with two children that are leaves with labels  $\fchild{p_i} = out$ and $\tchild{p_i} = out$ (\Cref{fig:hardness-initial_state-first}). Hence, initially, any datapoint will receive the label $out$ from each rule $\rl_i$. Furthermore, the is an additional rule $\rl_{xin}$ with a single predicate $p_{xin}: \exists i,j: v = a_{i,j}$ with $\tchild{p_{xin}} = in$ and $\fchild{p_{xin}} = out$.
\item
The expected labels \complaints are:
  \[
    \complaints(\x) =
    \begin{cases}
      in &\text{if}\, \x \in U\\
      out &\text{if}\, \exists i: \x = b_i\\
      out &\text{if}\, \exists i,j: \x = a_{ij}\\
    \end{cases}
  \]
\item The space of predicates is $\preds = \{ (\var = \var) \}$, i.e., a single predicate that returns $\ltrue$ on all inputs.
\item The thresholds are set as follows:
  \begin{itemize}
  \item $\inat = \frac{1}{\card{\xcomplaints}}$, that is each datapoint has to receive a correct label by at least one of the rules.
  \item $\lat = \frac{M}{n + (n \cdot m^2) + M}$
   \item $\inat = 1$, i.e., no rule can abstain on any datapoint.
  \end{itemize}
\end{itemize}

We claim that there exists a set cover of size $k$ or less iff there exists a rule repair $\rseq$ of  $\rules$ with a cost of less than or equal to $n \cdot m + k \cdot (n \cdot m)$.
Before proving this statement, we first state several properties that any solution to the instance of the rule repair problem defined above is guaranteed to fulfill. First off, observe that as all datapoints fulfill the predicate $v = v$, a refinement can only switch the label of every datapoint that fulfills (does not fulfill) the predicate of a rule $\rl_i$'s root node. Furthermore, observe that for each rule $\rl_i$, the left child of the root cannot be refined to assign a different label to any datapoint not fulfilling the root's predicate $v \in S_i \lor \exists j: v = a_{ij}$ as in particular every $b_i$ does not fulfill the predicate and assigning the label $in$ to all $b_i$'s causes the rule to not fulfill the accuracy threshold $\laccuracy(\rl_i) \geq \lat$ as there are $M > n + n \cdot m^2$ datapoints in $\{ b_i \}$.
That is, for every set $S_i$ represented by rule $\rl_i$, we either have the choice to refine the right child of the root by replacing it with a predicate $v = v$ with a true child $in$ (and any label for the false child as no datapoint will end up at the false child). We will interpret refining a rule $\rl_i$ as including the set $S_i$ in the set cover. Refining rule $\rl_{xin}$ has cost $n\cdot m^2$ (all datapoints $a_{ij}$ receive an incorrect label) and, thus, no solution can refine this rule. In any solution, for each $e_i$, there has to be at least one refined rule $\rl_j$ for $e_i \in S_j$ to ensure that the $\iaccuracy(e_i) \geq \frac{1}{m+1}$ (this datapoint receives at least one $in$ label). All other datapoints already fulfill $iaccuracy$ through rule $\rl_{xin}$ that, as explained before, cannot be refined in any solution. In summary, any solution for the rule repair problem encodes a set cover that includes every $S_i$ such that $\rl_i$ got refined.

\proofpar{$\Rightarrow$}
Assume that there exists a set cover $S_{i_1}$, \ldots, $S_{i_k}$ of size $k$. We have to show that there exists a minimal repair $\rseq$ of cost $\rcost(\rseq) \leq n \cdot m + k \cdot (n \cdot m)$. We construct this repair by refining $\rl_{i_j}$ for each $S_{i_j}$. As $S_{i_1}$, \ldots, $S_{i_k}$ is a set cover of size $k$, the repair cost of $\rseq$ is $\sum_{j=1}^{k} \card{S_{i_j}} + n \cdot m = \left(\sum_{j=1}^{k} \card{S_{i_j}}\right) + k \cdot (n \cdot m) \leq n \cdot m + k \cdot (n \cdot m)$, because for each refined rule $\rl_i$, the output of $\rl_{i}$ on all datapoints $e_j \in S_i$ and all datapoints $a_{ij}$ for $\forall j \in [1,n\cdot m]$ are changed resulting in a cost of $\card{S_i} + n \cdot m$ per refined rule and the claimed cost for $\rcost(\rseq)$.

\proofpar{$\Leftarrow$} Assume that there exists a rule repair $\rseq$ of cost
less than or equal to $n \cdot m + k \cdot (n \cdot m)$. Recall that for each
$S_i$ there exist $n \cdot m$ datapoints $a_{ij}$ that fulfill the root
predicate of rule $\rl_i$. That is, if $\rl_i$ is refined in a solution then all
there are $n \cdot m + \card{S_i}$ datapoints (all $a_{ij}$ for
$j \in [1,n \cdot m]$ and $e_j$ for all $e_j \in S_i$) whose label is updated.
As $\sum_{j: \rl_j \text{ was refined}} \card{S_j} \leq n \cdot m$,
$\rcost(\rseq) \leq n \cdot m + k \cdot (n \cdot m)$ implies that at most $k$
rules are refined by $\rseq$ and, thus,
$\card{\{ S_i \mid \rl_i \text{ was refined by } \rseq\}} \leq k$ and
$\{ S_i \mid \rl_i \text{ was refined by } \rseq\}$ is a set cover of size
$\leq k$.
\end{proof}

 \Cref{theo:rule-repair-hardness} states that the rule repair problem is \nphard. However, as we demonstrate experimentally, for rule sets of typical size we can determine exactly how to change the outputs of rules such that rules with this updated output are a solution to the rule repair problem in \cref{sec:repair}. Then we can apply any one of our single rule repair algorithms to determine a refinement repair sequence \rseq such that the updated rules  $\rseq(\rules)$ return the desired outcome on \xcomplaints. This is guaranteed to succeed based on \Cref{lem:bounded-repair-cost} that shows that as long as the predicate space is partitioning, then any change to a rule's output on a set of datapoints can be achieved by a refinement repair. However, achieving such an update with a minimal number of refinement steps (new predicates added to the rules) is computationally hard as the following theorem shows.


\begin{theorem}[Minimal Single Rules Refinement Repairs]\label{theo:minimal-refinement-r}
  Consider a rule \rl and a set of datapoints $\xcomplaints$ and labels $\complaints: \xcomplaints \to \outdb$.
  The following problem is \nphard:
  \[
    \argmin_{\rseq: \forall \x \in \complaints: \rseq(\rl)(\x) = \complaints(\x)} \pcost(\rseq)
  \]
\end{theorem}
\begin{proof}

  We prove this theorem by reduction from the \npcomplete Set Cover problem. Recall the set cover problem is given a set $U = \{e_1, \ldots, e_n\}$ and subsets $S_1$ to $S_m$ such that $S_i \subseteq U$ for each $i$, does there exist a $i_1$, \ldots, $i_k$ such that $\bigcup_{j=1}^{k} S_{i_j} = U$. Based on an instance of the set cover problem, we construct an instance of the rule repair problem as follows:
\begin{itemize}
\item The instance $\xcomplaints$ has  $n+1$ datapoints
$\{e_1, \ldots, e_n, b\}$ where $b \not\in U$.
\item Labels $\outdb = \{  out, in \}$
\item Rules $\rules = \{\rl \}$ where $\rl$ has a single predicate $p: (v = v)$ (i.e., it corresponds to truth value $true$) with two children that are leaves with labels  $\fchild{p} = in$ and $\tchild{p} = out$ (\Cref{fig:hardness-initial_state}). Hence, initially, any assignment will end up in $\tchild{p} = out$.
\item
The ground truth labels \complaints assigns a label to every datapoint as shown below.
  \[
    \complaints(e) =
    \begin{cases}
      in &\text{if}\, e \neq b\\
      out &\text{otherwise} (e = b)\\
    \end{cases}
  \]
\item Furthermore, the model $\model_{\rules}$ is defined as $\model(\x) = \rl(\x)$ (the model returns the labels produced by the single rule $\rl$).
\item The space of predicates is $\preds = \{ \var \in S_i \mid i \in [1,m] \}$
\end{itemize}
We claim that there exists a set cover of size $k$ or less iff there exists a minimal repair of  $\rules$ with a cost of less than or equal to $k$.

\begin{figure}[h]
    \centering
    \begin{minipage}{\linewidth}
        \centering
        \begin{tikzpicture}[scale=.8,
          level 1/.style={level distance=1.0cm, sibling distance=15mm},
          level 2/.style={level distance=1.0cm, sibling distance=15mm},
          level 3/.style={level distance=0.8cm, sibling distance=5mm},
          level 4/.style={level distance=0.8cm, sibling distance=5mm}]
          
          \node[logic](a) {$v \in S_i \lor \exists j: v = x_{ij}$}
              child {node[snode](b) {out} 
                edge from parent node[left] {\small \lfalse}}
              child {node[snode](c) {out} 
                edge from parent node[right] {\small \ltrue}};
              
        \end{tikzpicture}
    \end{minipage}
    \caption{Rule $\rl_i$ used in \Cref{proof:rule set repair hardness}}
    \label{fig:hardness-initial_state-first}
\end{figure}

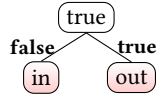
\begin{figure}[htb]
\begin{minipage}{\linewidth}
  \centering
    \begin{tikzpicture}[scale=.8,
      level 1/.style={level distance=1.0cm, sibling distance=15mm},
      level 2/.style={level distance=1.0cm, sibling distance=15mm},
      level 3/.style={level distance=0.8cm, sibling distance=5mm},
      level 4/.style={level distance=0.8cm, sibling distance=5mm}]
       \node[logic](a) {true}
      child {node[snode](b) {in} 
        edge from parent node[left] {\small\lfalse}}
      child {node[snode](c) {out} 
        edge from parent node[right] {\small\ltrue}};
    \end{tikzpicture}\\
    \vspace{-4mm}
    \caption{The rule representation of rule r used in \Cref{theo:minimal-refinement-r}}
    \label{fig:hardness-initial_state}
    \end{minipage}

    \end{figure}

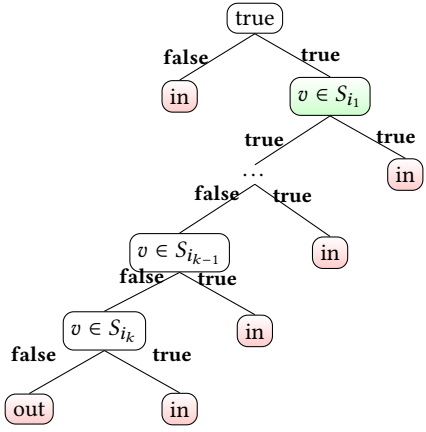
\begin{figure}[H]
    \centering
    \begin{minipage}{\linewidth}
    \centering
    \begin{tikzpicture}[scale=.8,
    level 1/.style={level distance=1.3cm, sibling distance=2.5cm},
    level 2/.style={level distance=1.3cm,sibling distance=2.5cm},
    level 3/.style={level distance=1.3cm, sibling distance=2.5cm},
    level 4/.style={level distance=1.3cm, sibling distance=2.5cm},
     level 5/.style={level distance=1.3cm, sibling distance=2.5cm}]
     \node[logic](a){true}
        child { node[snode](b) {in} 
            edge from parent node[rectangle, left] {\small\lfalse}
        }
        child { node[newlogic](c) {$v \in S_{i_1}$}
            child { node[draw=none](d) {\ldots}
                child { node[logic](e) {$v \in S_{i_{k-1}}$}
                    child { node[logic](f) {$v \in S_{i_{k}}$} 
                        child { node[snode](g) {out} edge from parent node[above left=1pt] {\small\lfalse} }
                        child { node[snode](h) {in} edge from parent node[above right=1pt] {\small\ltrue} }
                        edge from parent node[above] {\small\lfalse}
                    }
                    child { node[snode](i) {in} edge from parent node[above] {\small\ltrue} }
                    edge from parent node[above] {\small\lfalse}
                }
                child { node[snode](j) {in} edge from parent node[above] {\small\ltrue} }
                edge from parent node[left] {\small\ltrue}
            }
            child { node[snode](k) {in}
                edge from parent node[right] {\small\ltrue}
            }
            edge from parent node[right] {\small\ltrue}
        };
    \end{tikzpicture}
    \end{minipage}

    \caption{The rule representation of repair of rule r in  \Cref{proof:rule set repair hardness}} \label{fig:proof-np-hardness}
    \end{figure}

\proofpar{(only if)} 
Let $S_{i_1}, \ldots S_{i_k}$ be a set cover. We have to show that there exists a minimal repair $\rseq$ of size $\leq k$. We construct that repair as follows: we replace the \ltrue child of the single predicate $p: (\var = \var)$ in $\rl$ with a left deep tree with predicates $p_{i_j}$ for $j \in [1,k]$ such that $p_{i_j}$ is $(\var \in S_{i_j})$ and $\fchild{p_{i_j}} = p_{i_{j+1}}$ unless $j = k$ in which case $\fchild{p_{i_j}} = out$; for all $j$, $\tchild{p_{i_j}} = in$. The rule tree  for this rule is shown in \Cref{fig:proof-np-hardness}
The repair sequence $\rseq = \rop_1, \ldots, \rop_k$ has cost $k$. Here $\rop_i$ denotes the operation that introduces $p_{i_j}$ .

It remains to be shown that $\rseq$ is a valid repair with accuracy 1. Let $\rl_{up} = \rseq(\rl)$, we have $\rl_{up}(e) =
\complaints(e)$ for all $(e) \in R$. Recall that
$\model(e) = \rl(e)$ and, thus, $\rl_{up}(e) =
\complaints(e)$ ensures that $\model(e) = \complaints(e)$.
Since the model corresponds a single rule, we want $\rl_{up}(e_i) = in$ for all $i = 1, \cdots, n$, and $\rl_{up}(b) = out$.
Note that the final label is $out$ only if the check $(\var \in S_{i_j})$ is false for all $j = 1, \cdots, k$.
Since $S_{i_1}, \ldots S_{i_k}$ constitute a set cover, for every $e_i$, $i = 1, \cdots, n$, the check will be true for at least one $S_{i_j}$, resulting in a final label of $in$. On the other hand, the check will be false for all $S_{i_j}$ for $b$, and therefore the final label will be $out$. This gives a refined rule with accuracy 1.

\proofpar{(if)}
Let $\rseq$ be a minimal repair of size $\leq k$ giving accuracy 1. Let $\rl_{up} = \rseq(\rl)$, i.e., in the repaired rule $\rl_{up}$, all $e_i$, $i = 1, \cdots, n$ get the $in$ label and $b$ gets $out$ label.
We have to show that there exists a set cover of size $\leq k$.  First, consider the path taken by element $b$.
For any predicate $\pred \in \preds$ of the form $(v \in S_i)$, we have $\pred(b) = \lfalse$.
Let us consider the path
$\apath = \node_{root} \xrightarrow{b_1} \node_1 \xrightarrow{b_2} \node_2 \ldots \node_{l-1} \xrightarrow{b_{l}} \node_{l}$  taken by $b$.

As $\node_{root} = \pred_{root} = true$, we know that $b_1 = \ltrue$ ($b$ takes the $\ltrue$ edge of $\node_{root}$. Furthermore, as $\pred(b)$ for all predicates in $\preds$ (as $b \not\in U$),  $b$ follows the $\lfalse$ edge for all remaining predicates on the path.
That is $b_i = \lfalse$ for $i > 1$. As we have $\complaints(b) = out$ and $\rseq$ is a repair, we know that $\node_{l} = out$. For each element $e \in U$, we know that $\rl_{up}(e) = in$ which implies that the path for $e$ contains at least one predicate $\var \in S_i$ for which $e \in S_i$ evaluates to true.
To see why this has to be the case, note that otherwise $e$ would take the same path as $b$ and we have $\rl_{up}(e) = out \neq in = \complaints(e)$ contradicting the fact that $\rseq$ is a repair. That is, for each $e \in U$ there exists $S_i$ such $e \in S_i$ and $\pred_{i}: \var \in S_i$ appears in the tree of $\rl_{up}$. Thus, $\{ S_i \mid \pred_i \in \rl_{up}\}$ is a set cover of size $\leq k$.
\end{proof}

\cut{
===================================================\\
To prove the theorem, we first define a proxy problem, called the {\em Formula Change Problem}: given a 3CNF formula $\phi$ with $m$ clauses, an assignment $\alpha$, and an integer $k < m$, decide whether $\phi$ can be changed to $\phi'$ by adding $\leq k$ literals to its clauses, such that $\alpha(\phi) = True$. This problem is \nphard:

\begin{lemma}\label{lem:formula-change-hardness}
The Formula Change Problem is NP-hard.
\end{lemma}
\begin{proof}
By reduction from Max-3SAT. Max-3SAT is the problem of deciding whether at least $k$ clauses of a given 3CNF formula can be satisfied.
Given a datapoint of Max-3SAT $(\phi,k)$, where $\phi$ is a 3CNF formula and $k$ is an integer, we define a datapoint of the Formula Change Problem as follows: $\phi$ is the formula, $k' = m-k$, where $m$ is the number of clauses in $\phi$, and $\alpha$ is an arbitrary assignment.
We now claim that $\alpha$ satisfies at least $k$ clauses of $\phi$ iff $\phi$ can be changed to a formula $\phi'$ by adding at most $k'$ literals to $\phi$ such that $\alpha(\phi') = True$.

\proofpar{If direction}
Assume that $\alpha$ satisfies at least $k$ clauses of $\phi$. Add to each of the unsatisfied clauses $C_i$ a literal $l_j$ from $\phi$ such that $\alpha(l_j) = True$ such that the new clause is $C_i' = \bigvee_{p=1}^t l_p \lor l_j$. Call the new formula $\phi'$. Thus, all the clauses that were satisfied by $\alpha$ in $\phi$ remain satisfied in $\phi'$ and the rest of the clauses are now satisfied by the addition of $l_j$.

\proofpar{Only if direction}
Assume that $\phi$ can be changed to a formula $\phi'$ by adding at most $k'$ literals to $\phi$ such that $\alpha(\phi') = True$. The change from $\phi$ to $\phi'$ affects at most $k' = m-k$ clauses. Thus, at least $k$ of the clauses remain as they were in $\phi$. Therefore, these clauses are satisfied in $\phi$ as well, so $\alpha$ satisfies at least $k$ clauses in $\phi$.
\end{proof}

\begin{figure}[htb]
    \centering
    \begin{tikzpicture}[scale=.8,
    level 1/.style={level distance=.6cm, sibling distance=10mm},
    level 2/.style={sibling distance=13mm}, level distance=0.6cm,
    level 3/.style={level distance=0.6cm, sibling distance=13mm},
    level 4/.style={level distance=0.2cm, sibling distance=-10mm}]
    \Tree
    [.\node[logic] {$C_i[x_1]$};
    [.\node[logic] {$C_i[x_2]$};
    [.\node[logic] {$C_i[\overline{x_3}]$};
    \node[snode] {$\lfalse$}; \node[snode] {$\ltrue$};
    ]
    \node[snode] {$\ltrue$};
    ]
    \node[snode] {$\ltrue$};
    ]
    \end{tikzpicture}
    \caption{The rule representation of a 3CNF conjunct $C_i = (x_1 \lor x_2 \lor \overline{x_3})$ for the proof of \Cref{theo:rule-repair-hardness}} \label{fig:conjunct-rule}
\end{figure}

\begin{proof}[Proof of \Cref{theo:rule-repair-hardness}]
We prove this theorem by reduction from the Formula Change Problem.

Denote the 3CNF formula by $\phi = \bigwedge_{i=1}^m C_i$, where $C_i$ is a disjunction of three literals.
First, it is clear that every $C_i$ can be converted to a rule representation as described in \Cref{def:rule} in polynomial time and its representation includes $3$ inner nodes and $4$ label leaves ($3$ of them $True$ and $1$ is $False$) as demonstrated in \Cref{fig:conjunct-rule}.
We, therefore, denote the rule representation of $C_i$ as $\rl_{C_i}$.

Now for the reduction: given an datapoint of the Formula Change Problem $(\phi, \alpha, k)$, we define an datapoint of the rule repair problem as follows. $\rules = \{\rl_{C_1}, \ldots, \rl_{C_m}\}$, $\indb = \{\alpha\}$, $\outdb = \{True, False\}$, $\tau = 1$, $k' = k$, $C = \{(\alpha, True)\}$, and $\model_{\rules} = \bigwedge_{i=1}^m x(\rl_{C_i})$,
i.e., the conjunction of the rule labels given to the assignment (which acts as the datapoint). This translates to a Boolean evaluation of all rules for a given assignment ($\alpha(\phi)$ in our case).
We now claim that $\phi$ can be changed to a formula $\phi'$ by adding at most $k$ literals to $\phi$ such that $\alpha(\phi') = True$ iff $\phi$ can be repaired using at most $k$ operations from \Cref{sec:repair-operations} such that the label of $\alpha$ is $True$ ($\model_{\rules'}(x)=y_x$).

(If direction) Assume that $\phi$ can be changed to a formula $\phi'$ by adding at most $k$ literals to $\phi$ such that $\alpha(\phi') = True$. At most $k$ clauses could have been changed, so for every unchanged clause $C_i$, we know that $\alpha(C_i) = True$, and thus $\alpha(\rl_{C_i}) = True$.
Notice that adding a literal to $\phi$ is equivalent to adding a node to the rule representation of a clause in $\phi$ (as in described in\Cref{sec:repair-operations}).
Denote the augmented rules by $\rl_{C_i}'$, so $\alpha(\rl_{C_i}') = True$.
Hence, after the repair, $\alpha$ satisfies the augmented rules and the rules that remain unchanged, and thus, $\model_{\{\rules'\}}(\alpha)=\bigwedge_{i=1}^m \alpha(\rl_{C_i})=True$.

(Only if direction) Assume that the rules $\rules$ can be repaired using most $k$ operations from \Cref{sec:repair-operations} such that the label of $\alpha$ is $True$.
A rule $\rl_{C_i}$ can either be deleted or changed by a node addition at the end of the only path in $\rl_{C_i}$ that leads to the label $False$ (see  \Cref{fig:conjunct-rule} for an example).
Again, denote the augmented rules by $\rl_{C_i}'$, so $\alpha(\rl_{C_i}') = True$ or $\rl_{C_i}'$ has been deleted.
We change $\phi$ to $\phi'$ as follows. If $\rl_{C_i}$ was change to $\rl_{C_i}'$ by adding a literal at the end of the only path leading to $False$, we add this literal to the dijunction of $C_i$. Otherwise, $\rl_{C_i}$ was deleted, so we change $C_i$ by adding to its disjunction a literal $l_j$ from one of the remaining clauses (there are at least $m-k$ remaining clauses) such that $\alpha(l_i) = True$ to obtain $\phi'$. Thus, $\alpha(\phi') = True$.
\end{proof}
}


\section{Single Rule Refinement - Proofs and Additional Details}
\label{sec:single-rule-refin}

\subsection{Independence of Path Repairs}\label{sec:proof-path-indendence}

As refinements only extend existing paths in $\rl$ by replacing leaf nodes with new predicate nodes, in any refinement $\rl'$ of $\rl$, the path for a datapoint $\x \in \cdp$ has as prefix a path from $\fixp$. That is, for $\apath_1 \neq \apath_2 \in \fixp$, any refinement of $\apath_1$ can only affect the labels for datapoints in $\pdps{\apath_1}$, but not the labels of datapoints in $\pdps{\apath_2}$ as all datapoints in $\pdps{\apath_2}$ are bound to take paths in any refinement $\rl'$ that start with $\apath_2$. Hence we can determine repairs for each path in $\fixp$ independently.

\subsection{Partitioning Predicate Spaces}
\label{sec:part-pred-spac}

\begin{lemma}\label{lem:predicate-partitioning}
  Consider a space of predicates $\preds$ and atomic units $\atomicunit$.
\begin{itemize}
\item \textbf{Atomic unit comparisons}: If $\preds$ contains for every $\au \in \atomicunit$, constant $c$, and variable $\var$, the predicate $\ac{\var}{\au} = c$, then $\preds$ is \emph{partitioning}.
\item \textbf{Labeling Functions}: Consider the document labeling usecase. If $\preds$ contains predicate $w \in \var$ every word $w$, then $\preds$ is partitioning.
\end{itemize}
\end{lemma}
\begin{proof}
\paratitle{Atomic unit comparisons}
  Consider an arbitrary pair of datapoints $\x_1 \neq \x_2$ for some rule $\rl$ over $\preds$. Since, $\x_1 \neq \x_2$ it follows 
that  there has to exist $\au \in \atomicunit$ such that $\aa{\x_1}[\au] = c \neq \aa{\x_2}[\au]$. Consider the predicate $p = (\aa{\var}[\au] = c)$ which based on our assumption is in $\preds$.
  \begin{align*}
    (\ac{\x_1}{\au} = c) = \ltrue \neq \lfalse =  (\ac{\x_2}{\au} = c)  \\
  \end{align*}


\paratitle{Document Labeling}
Recall that the atomic units for the text labeling usecase are words in a sentence. Thus, the claim follows from the atomic unit comparisons claim proven above.

\end{proof}

\subsection{Existence of Path Repairs}
\label{sec:proof-crefl-repa}

Next, we will show that is always possible to find a refinement repair for a path if the space of predicates \preds is \emph{partitioning}, i.e., if 
for any two datapoints $\x_1 \neq \x_2$ there exists $\pred \in \preds$ such that:
  \[
    \pred(\x_1) \neq \pred(\x_2)
  \]
  Observe that any two datapoints $\x_1 \neq \x_2$ have to differ in at least one atomic unit, say $A$:
  $x_1[A] = c \neq x_2[A]$. 
  If $\preds$ includes all comparisons of the form $\ac{\var}{A} = c$, then any two datapoints can be distinguished. In particular, for labeling text documents, where the atomic units are words, \preds is partitioning if it contains $w \in \var$ for every word $w$.
The following proposition shows that when \preds is partitioning, we can always find a refinement repair. Further, we show an upper bound on the number of predicates to be added to a path $\apath \in \fixp$ to assign the ground truth labels $\pathc{\apath}$ to all datapoints in $\pathdps{\apath}$.

\begin{lemma}[Upper Bound on Repair Cost of a Path]\label{lem:bounded-repair-cost}
  Consider a rule \rl, a path \apath in \rl, a partitioning predicate space $\preds$, and ground truth labels $\pathc{\apath}$ for datapoints $\pathdps{\apath}$ on path \apath. Then there exists a refinement repair \rseq for path \apath and $\pathc{\apath}$ such that:
    $\rcost(\rseq) \leq \card{\pathdps{\apath}}$.

\end{lemma}
\begin{proof}
  We will use $y_{\x}$ to denote the expected label for $\x$, i.e., $\lab_{\x} = \asslabel(\x)$. Consider the following recursive greedy algorithm that assigns to each $\x \in \pathdps{\apath}$ the correct label. The algorithm starts with $\indb_{cur} = \pathdps{\apath}$ and in each step finds a predicate $\pred$ that ``separates'' two datapoints  $\x_1$ and $\x_2$ from $\indb_{cur}$ with $y_{\x_1} \neq y_{\x_2}$. That is, $\pred(\x_1)$ is true and $\pred(\x_2)$ is false. As $\preds$ is partitioning such a predicate has to exist. Let $\indb_1 = \{ \x \mid \x \in \indb_{cur} \land \pred(\x) \}$ and $\indb_2 = \{ \x \mid \x \in \indb_{cur} \land \neg \pred(\x) \}$. We know that $\x_1 \in \indb_1$ and $\x_2 \in \indb_2$. That means that $\card{\indb_1} < \card{\pathdps{\apath}}$ and $\card{\indb_2} < \card{\pathdps{\apath}}$.
  The algorithm repeats this process for $\indb_{cur} = \indb_1$ and $\indb_{cur} = \indb_2$ until all datapoints in $\indb_{cur}$ have the same desired label which is guaranteed to be the case if $\card{\indb_{cur}} = 1$. In this case, the leaf node for the current branch is assigned this label. As for each new predicate added by the algorithm the size of $\indb_{cur}$ is reduced by at least one, the algorithm will terminate after adding at most $\card{\pathdps{\apath}}$ predicates.
\end{proof}

\subsection{Non-minimality of the Algorithm from \Cref{lem:bounded-repair-cost}}
\label{sec:app-non-minimality-of-naive-greedy}
\begin{figure}[t]
  \begin{minipage}{0.49 \linewidth}
    \centering
    \footnotesize
    \begin{tikzpicture}[scale=.65,
    level 1/.style={level distance=1.5cm, sibling distance=30mm},
    level 2/.style={level distance=1.5cm,sibling distance=30mm},
    level 3/.style={level distance=1.5cm, sibling distance=30mm},
    level 4/.style={level distance=0.8cm, sibling distance=5mm},
    snode/.style = {shape=rectangle, rounded corners, draw, align=center, top color=white, bottom color=red!20},
    logic/.style = {shape=rectangle, rounded corners, draw, align=center, top color=toplogiccolor, bottom color=bottomlogiccolor},
    newlogic/.style = {shape=rectangle, rounded corners, draw, align=center, top color=white, bottom color=green!20},-]
    \node[logic] {$stars \in \var$}
      child {node[snode] {\abstain} 
        edge from parent node[above left] {{\small\lfalse}}
      }
      child {node[newlogic] {$four \in \var$} 
        child {node[newlogic] {$five \in \var$} 
          child {node[snode] {\negative} edge from parent node[above left] {{\small\lfalse}}}
          child {node[snode] {\positive} edge from parent node[above right] {{\small\ltrue}}}
          edge from parent node[above left] {{\small\lfalse}}}
        child {node[snode] {\positive} edge from parent node[above right] {{\small\ltrue}}}
        edge from parent node[above right] {{\small\ltrue}}
      };
    \end{tikzpicture}
  \end{minipage}
\begin{minipage}{0.49 \linewidth}
    \centering\footnotesize
	    \begin{tikzpicture}[scale=.65,
    level 1/.style={level distance=1.5cm, sibling distance=30mm},
    level 2/.style={level distance=1.5cm,sibling distance=30mm},
    level 3/.style={level distance=0.8cm, sibling distance=5mm},
    level 4/.style={level distance=0.8cm, sibling distance=5mm},
    snode/.style = {shape=rectangle, rounded corners, draw, align=center, top color=white, bottom color=red!20},
    logic/.style = {shape=rectangle, rounded corners, draw, align=center, top color=toplogiccolor, bottom color=bottomlogiccolor},
    newlogic/.style = {shape=rectangle, rounded corners, draw, align=center, top color=white, bottom color=green!20},-]]
    \node[logic] {$stars \in \var$}
      child {node[snode] {\abstain} edge from parent node[above left] {{\small\lfalse}}}
      child {node[newlogic] {$great \in \var$} 
         child {node[snode] {\negative} edge from parent node[above left] {{\small\lfalse}}}
         child {node[snode] {\positive} edge from parent node[above right] {{\small\ltrue}}}
         edge from parent node[above right] {{\small\ltrue}}
      };

    \end{tikzpicture}
  \end{minipage}

    \caption{A non-optimal rule repair for the LF from \Cref{ex:minimal-rule-repair-dc} produced by the algorithm from \Cref{lem:bounded-repair-cost} and an optimal repair (right) } \label{fig:non-optimal-dc-repair}
\end{figure}
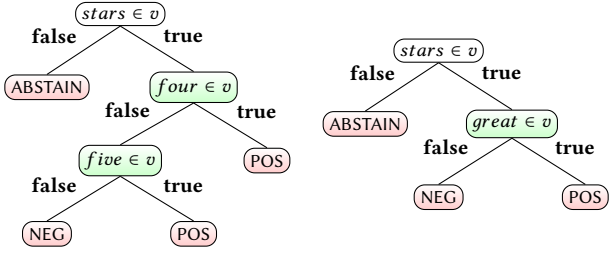
We demonstrate the non-minimality by providing an example on which the algorithm returns a non-minimal repair of a rule.

\begin{example}[]\label{ex:minimal-rule-repair-dc}
Consider \indb as shown below with 3 documents, their current labels (\negative) assigned by a rule and expected labels from \complaints. The original rule consists of a single predicate $stars \in \var$, assigning all documents that contain the word ``\emph{stars}'' the label \negative. The algorithm may repair the rule by first adding the predicate $four \in \var$ which separates $d_1$ and $d_3$ from $d_2$. Then an additional predicate has to be added to separate $d_1$ and $d_3$, e.g., $five \in \var$. The resulting rule is shown \Cref{fig:non-optimal-dc-repair} (left). The cost of this repair is $2$. However, a repair with a lower costs exists: adding the predicate $great \in \var$ instead. This repair has a cost of $1$. The resulting rule is shown in \Cref{fig:non-optimal-dc-repair} (right).


  \begin{center}
    {\footnotesize
\begin{tabular}{|l|c|c|}
  \hline
  \thead{sentence}                              & \thead{Current label} & \thead{Expected labels from \complaints} \\ \hline
  $d_1$: I rate this one stars. This is bad.    & \negative             & \negative                                \\
  $d_2$: I rate this four stars. This is great. & \negative             & \positive                                \\
  $d_3$: I rate this five stars. This is great. & \negative             & \positive                                \\
  \hline
\end{tabular}
}

\end{center}
\end{example}

\subsection{Equivalence of Predicates on Datapoints}
\label{sec:equiv-pred-assignm}

Our results stated above only guarantee that repairs with a bounded cost exist. However, the space of predicates may be quite large or even infinite. We now explore how equivalences of predicates wrt. $\pathdps{\apath}$ can be exploited to reduce the search space of predicates and present an algorithm that is exponential in $\card{\pathdps{\apath}}$ which determines an optimal repair independent of the size of the space of all possible predicates. First observe that with $\card{\pathdps{\apath}} = n$, there are exactly $2^n$ possible outcomes of applying a predicate \pred to $\pathdps{\apath}$ (returning either true or false for each $\x \in \pathdps{\apath}$). Thus, with respect to the task of repairing a rule through refinement to return the correct label for each $\x \in \pathdps{\apath}$, two predicates are equivalent if they return the same result on \assigns in the sense that in any repair using a predicate $\pred_1$, we can substitute $\pred_1$ for a predicate $\pred_2$ with the same outcome and get a repair with the same cost. That implies that when searching for optimal repairs it is sufficient to consider one predicate from each equivalence class of predicates.

\begin{lemma}[Equivalence of Predicates]\label{lem:equivalence-of-predicates}
  Consider a space of predicates \preds, rule \rl, and set of datapoints \pathdps{\apath} with associated expected labels \pathc{\apath} and assume the existence of an algorithm $\mathcal{A}$ that computes an optimal repair for \rl given a space of predicates. Two predicates $\pred \neq \pred' \in \preds$ are considered equivalent wrt. \pathdps{\apath}, written as $\pred \equiv_{\pathdps{\apath}} \pred'$ if $\pred(\x) = \pred'(\x)$ for all $\x \in \pathdps{\apath}$. Furthermore, consider a reduced space of predicates $\preds_{\equiv}$ that fulfills the following condition:
  \begin{align}\label{eq:pred-equivalence-classes}
    \forall \pred \in \preds: \exists \pred' \in \preds_{\equiv}: \pred \equiv \pred'
  \end{align}
  For any such $\preds_{\equiv}$ we have:
  \begin{align*}
    \rcost(\mathcal{A}(\preds, \rl, \pathc{\apath})) = \rcost(\mathcal{A}(\preds_{\equiv}, \rl, \pathc{\apath}))
  \end{align*}

\end{lemma}
\begin{proof}
Let $\rseq = \mathcal{A}(\preds_{\equiv}, \rl, \pathc{\apath})$ and $\rseq_{\equiv} = \mathcal{A}(\preds, \rl, \pathc{\apath}))$. Based on the assumption about $\mathcal{A}$, $\rseq$ ($\rseq_{\equiv}$) are optimal repairs within $\preds$ ($\preds_{\equiv}$).  We prove the lemma by contradiction. Assume that $\rcost(\rseq_{\equiv}) > \rcost(\rseq)$. We will construct from $\rseq$ a repair $\rseq'$ with same cost as $\rseq$ which only uses predicates from $\preds_{\equiv}$. This repair then has cost $\rcost(\rseq') = \rcost(\rseq) < \rcost(\rseq_{\equiv})$ contradicting the fact that $\rseq_{\equiv}$ is optimal among repairs from $\preds_{\equiv}$. $\rseq'$ is constructed by replacing each predicate $\pred \in \preds$ used in the repair with an equivalent predicate from $\preds_{\equiv}$. Note that such a predicate has to exist based on the requirement in \Cref{eq:pred-equivalence-classes}. As equivalent predicates produce the same result on every $\x \in \pathdps{\apath}$, $\rseq'$ is indeed a repair. Furthermore, substituting predicates does not change the cost of the repair and, thus, $\rcost(\rseq') = \rcost(\rseq)$.
\end{proof}

If the semantics of the predicates in $\preds$ is known, then we can further reduce the search space for predicates by exploiting these semantics and efficiently determine a viable $\pred_{\equiv}$. For instance, predicates of the form $A = c$ for a given atomic element $A$ only have linearly many outcomes on $\pathc{\apath}$ and the set of $\{\var.A = c\}$ for all atomic units $A$, variables in $\rules$, and constants $c$ that appear in at least one datapoint $\x \in \indb$ contains one representative of each equivalence class.\footnote{With the exception of the class of predicates that return false on all $\x \in \pathdps{\apath}$. However, this class of predicates will never be part of an optimal repair as it does only trivially partitions $\pathdps{\apath}$ into two sets $\pathdps{\apath}$ and $\emptyset$.}





}
\section{Path Refinement Repairs - Proofs and Additional Details}
\label{sec:path-refin-repa}

\subsection{\abbrGreedyPredRepair}
\label{sec:abbrgreedypredrepair}

\begin{algorithm}[h]
  \SetKwInOut{Input}{Input}\SetKwInOut{Output}{Output}
  \LinesNumbered
  \Input{Rule \rl\\
    Path $P$\\
    datapoints to fix $\pathdps{\apath}$\\
    Expected labels for assignments $\pathc{\apath}$\\
  }
  \Output{Repair sequence $\rseq$ which fixes $\rl$ wrt. $\pathc{\apath}$}
  \BlankLine
  $todo \gets [(\apath,\pathc{\apath})]$\\
  $\rseq = []$\\
  \While{$todo \neq \emptyset$}
  {\label{l:rr-todo-iteration}
    $(\apath,\pathc{\apath}) \gets pop(todo)$\\
    \If{$\exists \x_1, \x_2 \in \indb: \pathc{\apath}[\x_1] \neq \pathc{\apath}[\x_2]$}
    {\label{l:rr-pick-two}
      \tcc{Determine predicates that distinguish assignments that should receive different labels for a path}
      $p \gets \fgetsepp(\x_1,\x_2)$\label{l:rr-sep-pred}\\
      $y_1 \gets \pathc{\apath}[\x_1]$\\
      $\rop \gets \rrefine{\rl_{cur}}{\apath}{y_1}{\pred}{\ltrue}$\\
      $\indb_1 \gets \{ \x \mid \x \in \pathdps{\apath}\land \pred(\x) \}$\\
      $\indb_2 \gets \{ \x \mid \x \in \pathdps{\apath}\land \neg \pred(\x) \}$\\
      $todo.push((\rpath{\rl_{cur}}{\x_1}, \indb_1))$\label{l:rr-push-l1}\\
      $todo.push((\rpath{\rl_{cur}}{\x_2}, \indb_2))$\label{l:rr-push-l2}\\
    }
    \Else
    {\label{l:rr-pick-one}
      $\rop \gets \rrefinelabel{\rl_{cur}}{\apath}{\pathc{\apath}[\x]}$\\
    }
  $\rl_{cur} \gets \rop(\rl_{cur})$\\
  $\rseq.append(\rop)$\\
  }
  \Return $\rseq$
  \caption{\abbrGreedyPredRepair}
  \label{algo:naive-predicate-repair}
\end{algorithm}

The function \abbrGreedyPredRepair is shown \Cref{algo:naive-predicate-repair}.
This algorithm
 maintains a list of pairs of paths and datapoints at these paths to be processed. This list is initialized with all datapoints \pdps{\apath} from \pz{\apath} and the path \apath provided as input to the algorithm. In each iteration, the algorithm picks two datapoints $\x_1$ and $\x_2$ from the current set and selects a predicate
$\pred$ such that $\pred(\x_1) \neq \pred(\x_2)$. It then refines the rule with $\pred$ and appends $\indb_1 = \{ \x \mid \x \in \pathdps{\apath} \land \pred(\x) \}$ and $\indb_2 = \{ \x \mid \x \in \pathdps{\apath} \land \neg\,\pred(\x) \}$ with their respective paths to the 
list.  As shown in the proof of \iftechreport{\Cref{lem:bounded-repair-cost}}\ifnottechreport{\cite{li2025refininglabelingfunctionslimited}}, this algorithm terminates after adding at most $\card{\pathdps{\apath}}$ new predicates.

To ensure that all datapoints ending in path $\apath$ get assigned the desired label based on $\pathc{\apath}$, we need to add predicates to the end of $\apath$ to ``reroute'' each datapoint to a leaf node with the desired label.
As mentioned above, this algorithm implements the approach from the proof of \iftechreport{\Cref{lem:bounded-repair-cost}}\ifnottechreport{\cite{li2025refininglabelingfunctionslimited}}: for a set of datapoints taking a path with prefix $\apath$ ending in a leaf node that is not pure (not all datapoints in the set have the same expected label), we pick a predicate that \emph{``separates''} the datapoints, i.e., that evaluate to true on one of the datapoints and false on the other.
Our algorithm applies this step until all leaf nodes are pure wrt. the datapoints from $\pathdps{\apath}$. For that, we maintain a queue of path and datapoint set pairs which tracks which combination of paths and datapoint sets still have to be fixed. This queue is initialized with $\apath$ and all datapoints for $\apath$ from $\pathdps{\apath}$.  The algorithm processes sets of datapoints until the todo queue is empty. In each iteration,
the algorithm greedily selects a pair of datapoints $\x_1$ and $\x_2$ ending in this path that should be assigned different labels (\cref{l:rr-pick-two}). It then calls method \fgetsepp (\cref{l:rr-sep-pred}) to determine a predicate $\pred$ which evaluates to true on $\x_1$ and false on $\x_2$ (or vice versa). If we extend path $\apath$ with $\pred$, then $\x_1$ will follow the \ltrue edge of $p$ and $\x_2$ will follow the \lfalse edge (or vice versa). This effectively partitions the set of datapoints for path $\apath$ into two sets $\indb_1$ and $\indb_2$ where $\indb_1$ contains $\x_1$ and $\indb_2$ contains $\x_2$. We then have to continue to refine the paths ending in the two children of $p$ wrt. these sets of datapoints.
This is ensured by adding these sets of datapoints with their new paths to the todo queue (\cref{l:rr-push-l1,l:rr-push-l2}).
If the current set of datapoints does not contain two datapoints with different labels, then we know that all remaining datapoints should receive the same label. The algorithm picks one of these datapoints $\x$ (\cref{l:rr-pick-one}) and changes the current leave node's label to $\pathc{\apath}(\x)$.

\paratitle{Generating Predicates}
The implementation of
\fgetcovpred is specific to the type of \abbrRBBM. We next present implementations of this procedure for weak supervised labeling that exploit the properties of these two application domains. However, note that, as we have shown in \iftechreport{\Cref{sec:proof-crefl-repa}}\ifnottechreport{\cite{li2025refininglabelingfunctionslimited}}, as long as the space of predicates for an application domain contains equality and inequality comparisons for the atomic elements of datapoints, it is always possible to generate a predicate for two datapoints such that only one of these two datapoints fulfills the predicate. 
The algorithm splits the datapoint set $\xcomplaints$ processed in the current iteration into two subsets, which each are strictly smaller than $\xcomplaints$. Thus, the algorithm is guaranteed to terminate and by construction assigns each datapoints $\x$ in \pathdps{\apath} its desired label $\pathc{\apath}(\x)$.

\begin{algorithm}[t]
  \SetKwInOut{Input}{Input}\SetKwInOut{Output}{Output}
  \LinesNumbered
  \Input{Rule \rl\\
    Path $P$\\
    Datapoints to fix $\pathdps{\apath}$\\
    Expected labels for datapoints $\pathc{\apath}$\\
  }
  \Output{Repair sequence $\rseq$ which fixes $\rl$ wrt. $\pathc{\apath}$}
  \BlankLine
  $todo \gets [(\rl,\emptyset)]$\\
  $\preds_{all} = \fgetallpreds(P,\pathdps{\apath},\pathc{\apath})$\\
  \While{$todo \neq \emptyset$}
  {\label{l:bf-todo-iteration}
    $(\rl_{cur},\rseq_{cur}) \gets pop(todo)$\\
    \ForEach{$\apath_{cur} \in \leafpaths{\rl_{cur}, P}$}
    {
      \ForEach{$\pred \in \preds_{all} - \preds_{\rl_{cur}}$}
      {
        \ForEach{$\lab_1 \in \outdb \land \lab_1 \neq \plast{\apath_{cur}}$}
        {
          $\rop_{new} \gets \rrefine{\rl_{cur}}{\apath_{cur}}{y_1}{\pred}{\ltrue}$\\
          $\rl_{new} \gets \rop_{new}(\rl_{cur})$\\
          $\rseq_{new} \gets \rseq_{cur}, \rop_{cur}$\\
          \If{$\iaccuracy(\rl_{new}, \pathc{\apath})  = 1$}
          {
            \Return $\rseq_{new}$
          }
          \Else
          {
            $todo.push((\rl_{new},\rseq_{new}))$
          }
        }
      }
    }
  }
  \caption{\abbrOptimalPathRepair}
  \label{algo:optimal-path-repair}
\end{algorithm}

\subsection{\abbrOptimalPathRepair}
\label{sec:abbr-opt-path-repair-algo}

The brute-force algorithm (\Cref{algo:optimal-path-repair})
is optimal, i.e., it returns a refinement of minimal cost (number of new predicates added).
This algorithm enumerates all possible refinement repairs for a path $\apath$. Each such repair corresponds to replacing the last element on $\apath$ with
some rule tree. We enumerate such trees in increasing order of their size and pick the smallest one that achieves perfect accuracy on \pathc{\apath} wrt. $\pz{\apath}$.
We first determine all predicates that can be used in the candidate repairs. \iftechreport{As argued in \Cref{sec:equiv-pred-assignm}}\ifnottechreport{As shown in \cite{li2025refininglabelingfunctionslimited}}, there are only finitely many distinct predicates (up to equivalence) for a given set \pathdps{\apath}.
We then process a queue of candidate rules, each paired with the repair sequence that generated the rule.
In each iteration, we process one rule from the queue and extend it in all possible ways by replacing one leaf node, 
and selecting the refined rule with minimum cost that satisfies all assignments.
As we generate subtrees in increasing size, \iftechreport{\Cref{lem:bounded-repair-cost} implies  implies}\ifnottechreport{as shown in \cite{li2025refininglabelingfunctionslimited},} the algorithm will terminate and its worst-case runtime is exponential in $n=\card{\pathdps{\apath}}$ as it may generate all subtrees of size up to $n$.

\iftechreport{
\subsection{Proof of \Cref{theo:path-repair-correct}}
\label{sec:proof-path-algo-correctness}

\begin{proof}{Proof of \Cref{theo:path-repair-correct}}
In the following let $n = \card{\pathdps{\apath}}$.

  \proofpar{\abbrGreedyPredRepair}
  As \abbrGreedyPredRepair does implement the algorithm from the proof of \Cref{lem:bounded-repair-cost}, it is guaranteed to terminate after at most $n$ steps and produce a repair that assign to each $\x$ the label $\pathc{\apath}(\x)$.

  \proofpar{\abbrOptimalPathRepair}
 The algorithm generates all possible trees build from predicates and leaf nodes in increasing order of their side. It terminates once a tree has been found that returns the correct labels on $\pathdps{\apath}$. As there has to exist a repair of size $n$ or less, the algorithm will eventually terminate.

  \proofpar{\abbrEntropyPathRepair}
  This algorithm greedily selects a predicate in each iteration that minimizes the Gini impurity score. The algorithm terminates when for every leaf node, the set of datapoints from $\pathdps{\apath}$ ending in this node has a unique label. That is, if the algorithm terminates, it returns a solution. It remains to be shown that the algorithm terminates for every possible input.  As it is always possible to find a separator predicate $\pred$ that splits a set of datapoints $\pathdps{\apath}$ into two subsets $\indb_1$ and $\indb_2$ with fewer predicates which have a lower Gini impurity score than splitting into $\indb_1 = \pathdps{\apath}$ and $\indb_2 = \emptyset$, the size of the datapoints that are being processed, strictly decrease in each step. Thus, the algorithm will, in the worst-case, terminate after adding $n$ predicates.
\end{proof}
}

\begin{figure*}[htbp]
    \centering
        \begin{minipage}{\linewidth}
            \centering
            \includegraphics[trim={0 60 0 90}, clip, scale=0.5]{experiments/legend_runtime.png}
            \vspace{-6mm} 
        \end{minipage}
        
    \begin{subfigure}[b]{\linewidth}
        \centering
        \begin{minipage}{0.33\linewidth}
            \centering
            \includegraphics[scale=0.5]{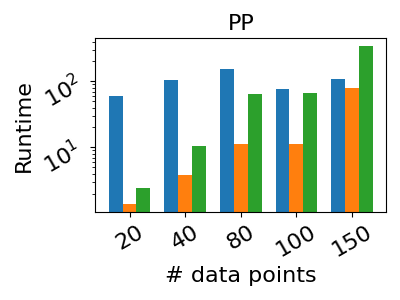}
        \end{minipage}
        \begin{minipage}{0.33\linewidth}
            \centering
            \includegraphics[scale=0.5]{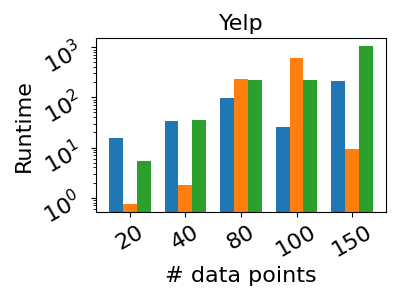}
        \end{minipage}
        \begin{minipage}{0.33\linewidth}
            \centering
            \includegraphics[scale=0.5]{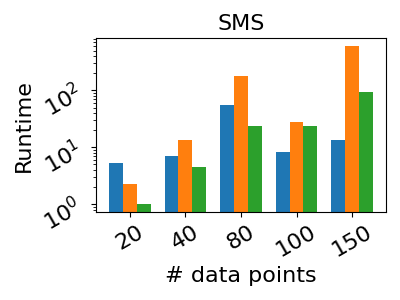}
        \end{minipage}
        \caption{Runtime, varying the size of \xcomplaints.}
        \label{exp_fig:additional_runtimes}
    \end{subfigure}


    \begin{subfigure}[b]{\linewidth}
        \centering\includegraphics[scale=0.35]{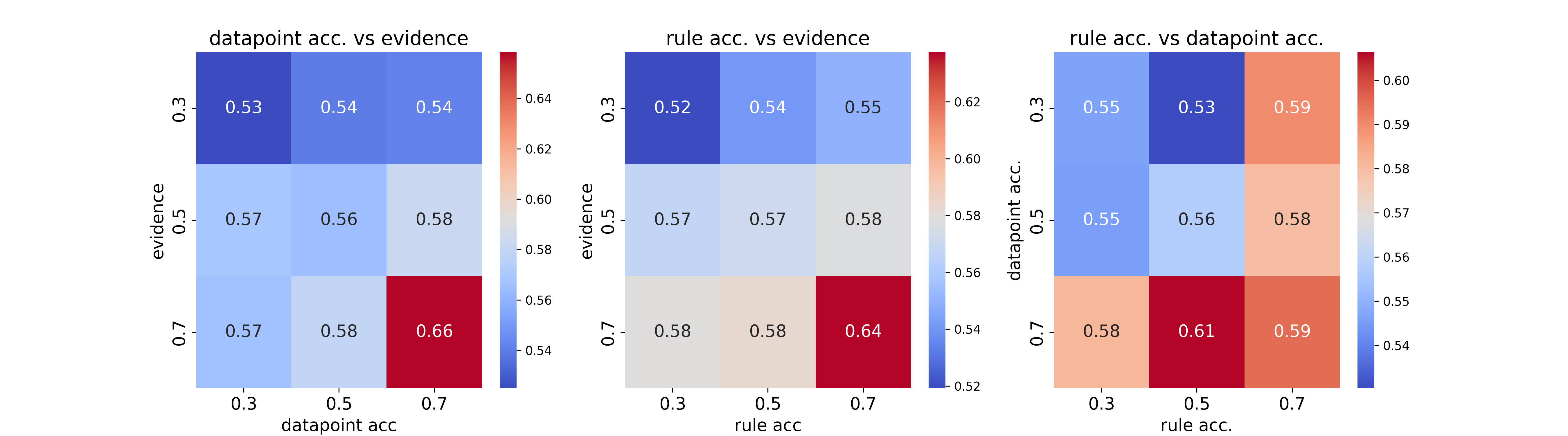}
        \caption{Pairwise interaction heatmaps for New Global Acc.}
        \label{exp_fig:heatmap}
    \end{subfigure}

    \caption{Additional experimental results}
    \label{fig:combined}
\end{figure*}

\begin{table}[htbp]
  \centering
  {\footnotesize
  \begin{tabular}{|c|r|r|c|c|c|c|}
\hline
\cellcolor{gray!40}\textbf{dataset}  & \cellcolor{gray!40}\textbf{repairer}  & \cellcolor{gray!40}\textbf{fix\%} &  \cellcolor{gray!40}\textbf{preserv\%} &  \cellcolor{gray!40}\textbf{global acc.} &  \cellcolor{gray!40}\textbf{new global acc.} \\
\hline
\DSfnews & RC  &   1 & 1	 & 0.71	& 0.92 \\
\hline
\DSfnews & LLM	 &   0.9 & 0.65	 & 0.71	& 0.81 \\
\hline
\DSamazon & RC  & 1 & 1 & 0.6 & 0.77 \\
\hline
\DSamazon & LLM  & 0.9 & 0.5 & 0.6 & 0.7 \\
\hline
  \end{tabular}
  }
  \caption{LLM vs \oursys (RC) quality rule refinement comparison}
  \label{tab:exp_gpt_vs_rc}
\end{table}



\section{Additional Experiment Details}
\subsection{runtime breakdown}\label{appendix:runtime}
The runtime breakdown of the rest of the datasets from \Cref{sec:exp-runtime} are shown in \Cref{exp_fig:additional_runtimes}.

\subsection{MILP thresholds}\label{appendix:milp-thresholds}
The effects on global accuracy of the pairwise relationships of $\iat$, $\inat$, and $\lat$ are shown in \Cref{exp_fig:heatmap}.
The color of a square represents global accuracy after the repair. Based on these results, it is generally preferable to set the thresholds higher as discussed in \Cref{sec:rule-repair-problem}. However, larger thresholds reduce the amount of viable solutions to the MILP and, thus, can significantly increase the runtime of solving the MILP and lead to overfitting to \xcomplaints.


\iftechreport{
\subsection{LLM as LF generator}\label{appendix:llm_lf_generator}

The prompt used to generate LFs mentioned in \Cref{sec:llm_vs_rc} is shown in \Cref{box:labeling}
}



\section{Use LLM as refinement}\label{append:llm-refine-detail}
In this section, we compare \oursys against a baseline using a large language model (LLM). We designed a prompt instructing the LLM to act as an assistant and refine the \glspl{lf} given a set of labeled datapoints. In this experiment, we used the \DSfnews dataset with 40 labeled datapoints and \DSamazon dataset with 20 labeled examples. We used \texttt{GPT-4-turbo} as the LLM.
A detailed description of the prompt and responses
from the LLM are presented in the \iftechreport{box below}\ifnottechreport{\cite{li2025refininglabelingfunctionslimited}}.



We manually inspected the rules returned by the LLM to ensure that they are semantically meaningful.
The quality of results after running Snorkel with the refined LFs from LLM and \oursys are shown in \Cref{tab:exp_gpt_vs_rc}. \texttt{fix\%} measures the percentage of the wrong predictions by Snorkel are fixed after retraining Snorkel with the refined rules. \texttt{preserv \%} measures what percentage of the input correct predictions by Snorkel remain the same after retraining with refined rules. \oursys outperforms the LLM in both global accuracy and accuracy on labeled input data. \iftechreport{Note that the purpose of comparing with LLMs is not to test if LLM can reproduce our algorithm but to test how good is LLM in performing the refinement task given the description. }We observe that the LLM tends preserve the semantic meaning of the original \glspl{lf} in the repairs it produces. For example, in one of the rules from \DSfnews, the original rule is \pyth{if 'talks' in text then REAL otherwise ABSTAIN} and the repaired rule was \pyth{if any(x in text for x in ['discussions', 'negotiations', 'talks']}. In one of the rules from \DSamazon, the original rule \pyth{if any (x in text for x in ['junk','disappointed','useless']) then NEGATIVE else ABSTAIN} mainly covers negative reviews. The LLM did add more negative words such as \pyth{'defective'} whereas \oursys could possibly add opposite sentiment conditions based on the solutions provided by the MILP. For example, it is possible for \oursys to refine a rule with negative sentiment by adding \pyth{else if 'great' in text then POSITIVE else ABSTAIN}. The returned refined functions for \DSfnews \iftechreport{are shown in \Cref{fig:code-part1,fig:code-part5,fig:amazon_llm_refined_code_1,fig:amazon_llm_refined_code_2,fig:amazon_llm_refined_code_3}}\ifnottechreport{can be found in \cite{li2025refininglabelingfunctionslimited}}.

\newpage
\iftechreport{
\begin{mdframed} 
You will help me refine a list of python functions that are being used on a classification task. I will provide you a list of function definitions and a dataset table.  The classification process is done in the following way: Given the dataset and the list of rules, there will be a black box aggregator that makes decisions by combining the results of the rules on the dataset. 

the dataset contains 2 columns: "text" and "expected label". You should apply the function to the given dataset and then modify the functions as you like. For this task, you can add new "if-else" statements or adjust the existing logic as you like.

return the refined labeling functions using the same formats as provided. Please add comments on why you changed the original function.
\end{mdframed}
}



\iftechreport{
\pagebreak
\begin{figure*}
    \centering
    \scriptsize
    \begin{python}
    def func1(text):
        # Expanded keyword coverage
        keywords = ["video", "featured", 
        "clip", "footage"]
        if any(x in text for x in keywords):
            return 0
        else:
            return -1
    def func2(text):
        # Expanded keyword coverage
        keywords1 = ["getty", "screenshot", "image"]
        keywords2 = ["registry", "database"]
        
        if any(x in text for x in keywords1):
            if any(x in text for x in keywords2):
                return 1
            else:
                return -1
        else:
            return -1

    def func3(text):
        # Expanded keyword coverage
        keywords = ["minister", "ministry", "parliament", 
                    "government", "official"]
        
        if any(x in text for x in keywords):
            return 0
        else:
            return -1

    def func4(text):
        # Expanded keyword coverage
        keywords = ["pic", "photo", "image", "snapshot"]
        
        if any(x in text for x in keywords):
            return 1
        else:
            return -1
            
    def func5(text):
        # Expanded keyword coverage
        keywords1 = ["wednesday", "spokesman", 
        "thursday", "representatives", 
        "nov", "tuesday", "monday"]
        keywords2 = ["legislation", "bill", 
        "law"]
        
        if any(x in text for x in keywords1):
            if any(x in text for x in keywords2):
                return 0
            else:
                return -1
        else:
            return -1

    def func6(text):
        # Expanded keyword coverage
        keywords = ["korea", "missile", "region",
                    "regional", "authorities",
                    "conflict", "war", "border"]
                    
        if any(x in text for x in keywords):
            return 0
        else:
            return -1
    \end{python}
    \caption{Refined \DSfnews funcs by LLM (Part 1)}
    \label{fig:code-part1}
\end{figure*}

\pagebreak

\begin{figure*}
    \centering
    \begin{python}
    def func7(text):
        # Expanded keyword coverage
        keywords1 = ["korea", "missile", "region",
                     "regional", "authorities", "conflict",
                     "war", "border"]
        keywords2 = ["korean", "diplomatic"]
        
        if any(x in text for x in keywords1):
            if any(x in text for x in keywords2):
                return 0
            else:
                return -1
        else:
            return -1

    def func8(text):
        # Expanded keyword coverage
        keywords = ["getty", "watch", "image",
                    "com", "https", "don", "woman",
                    "didn", "gop", "site", "web"]
        
        if any(x in text for x in keywords):
            return 1
        else:
            return -1

    def func9(text):
        keywords = ["talks", "negotiations", 
        "discussions"]
        
        if any(x in text for x in keywords):
            return 0
        else:
            return -1

    def func10(text):
        keywords = ["rex", "rex images", "rex news"]
        
        if any(x in text for x in keywords):
            return 1
        else:
            return -1

    def func11(text):
        # Expanded geopolitical terms
        keywords = ["northern", "turkey", 
                    "britain", "forces", "ruling", 
                    "european", "alliance", "nato", "eu"]
        
        if any(x in text for x in keywords):
            return 0
        else:
            return -1
    \end{python}
    \caption{Refined \DSfnews funcs by LLM (Part 2)}
    \label{fig:code-part5}
\end{figure*}

\pagebreak

\begin{figure*}
    \centering
    \begin{python}
    # func1: Improved to capture more 
    # variations of ratings, like "rating", 
    # "out of five", etc.
    def func1(text):
        keywords = ['star', 'stars', 
        'rating', 'out of five', 'rated']
        if any(x in text for x in keywords):
            return 0
        else:
            return -1

    # func2: Expanding product-related terms
    def func2(text):
        keywords = ['product', 'fit', 
        'quality', 'size', 'cheap', 'wear', 
        'material', 'design', 'durable']
        if any(x in text for x in keywords):
            return 1
        else:
            return -1

    # func3: No changes needed, it's quite 
    # simple but effective
    def func3(text):
        keywords = ['great']
        if any(x in text for x in keywords):
            return 0
        else:
            return -1

    # func4: Simplified to capture 
    # more positive combinations
    def func4(text):
        keywords1 = ['great', 'excellent', 'amazing']
        keywords2 = ['stars', 'works', 'fantastic', 
        'love']
        if any(x in text for x in keywords1):
            if any(x in text for x in keywords2):
                return 0
            else:
                return -1
        else:
            return -1

    # func5: Added more phrases for negative sentiment
    def func5(text):
        keywords = ['waste', 'not worth', 
        'disappointed', 'regret', 
        'poor value']
        if any(x in text for x in keywords):
            return 1
        else:
            return -1

    # func6: Added more keywords 
    # related to product 
    # discomfort or bad quality
    def func6(text):
        keywords = ['shoes', 'item', 'price',
        'comfortable', 'plastic',
        'uncomfortable', 'bad quality']
        if any(x in text for x in keywords):
            return 0
        else:
            return -1
    \end{python}
    \caption{Refined \DSamazon funcs by LLM (Part 1)}
    \label{fig:amazon_llm_refined_code_1}
\end{figure*}

\pagebreak

\begin{figure*}
    \centering
    \begin{python}
    # func7: Expanded list for 
    # disappointment and dissatisfaction
    def func7(text):
        keywords = ['junk', 'bought', 'like', 
        'dont', 'just', 'use', 'buy', 
        'work', 'small', 'didnt', 
        'did', 'disappointed', 'bad',
        'terrible', 'horrible', 'awful', 'useless']
        
        if any(x in text for x in keywords):
            return 1
        else:
            return -1

    # func8: Also improving dissatisfaction 
    # detection with more negative keywords
    def func8(text):
        keywords1 = ['junk', 'bought', 'like', 
        'dont', 'just', 'use', 'buy', 
        'work', 'small', 'didnt', 
        'did', 'disappointed', 'bad', 
        'terrible', 'horrible',
        'awful', 'useless']
        
        keywords2 = ['shoes', 'metal', 
        'fabric', 'replace', 'battery', 
        'warranty', 'plug', 'defective',
        'broken']
        if any(x in text for x in keywords1):
            if any(x in text for x in keywords2):
                return 1
            else:
                return -1
        else:
            return -1

    # func9: Added more positive words
    def func9(text):
        keywords = ['love', 'perfect', 
        'loved', 'nice', 'excellent',
        'works', 'loves', 'awesome', 
        'easy', 'fantastic', 'recommend']
        
        if any(x in text for x in keywords):
            return 0
        else:
            return -1

    # func10: Added more combination
    # cases for positive reviews
    def func10(text):
        keywords1 = ['love', 'perfect', 
        'loved', 'nice', 'excellent',
        'works', 'loves', 'awesome', 
        'easy', 'fantastic', 'recommend']
        
        keywords2 = ['stars', 'soft', 
        'amazing', 'beautiful']
        
        if any(x in text for x in keywords1):
            if any(x in text for x in keywords2):
                return 0
            else:
                return -1
        else:
            return -1
    \end{python}
    \caption{Refined \DSamazon funcs by LLM (Part 2)}
    \label{fig:amazon_llm_refined_code_2}
\end{figure*}

\pagebreak

\begin{figure*}
    \centering
    \begin{python}
    # func11: Adding more 
    # product-related terms 
    # and combinations
    def func11(text):
        keywords1 = ['love', 'perfect', 
        'loved', 'nice', 'excellent', 
        'works', 'loves', 'awesome',
        'easy', 'fantastic', 'recommend']
        
        keywords2 = ['shoes', 'bought', 
        'use', 'purchase', 'purchased',
        'colors', 'install', 'clean',
        'design', 'pair', 'screen', 
        'comfortable', 'products', 'item']
        
        if any(x in text for x in keywords1):
            if any(x in text for x in keywords2):
                return 0
            else:
                return -1
        else:
            return -1

    # func12: No change; this already targets 
    # product-related complaints well.
    def func12(text):
        keywords = ['returned', 'broke', 
        'battery', 'cable', 'fits', 
        'install', 'sturdy', 'ordered',
        'usb', 'replacement', 'brand',
        'installed', 'unit', 'batteries', 
        'box', 'warranty', 'defective', 
        'cheaply', 'durable', 'advertised']
        
        if any(x in text for x in keywords):
            return 1
        else:
            return -1

    # func13: Adding a few more 
    # fun product terms for cuteness
    def func13(text):
        keywords = ['cute', 'shirt', 
        'adorable', 'lovely', 'sweet']
        
        if any(x in text for x in keywords):
            return 0
        else:
            return -1

    # func14: Improved negative terms 
    # related to fabric and poor quality
    def func14(text):
        keywords = ['fabric', 'return',
        'money', 'poor', 'garbage', 
        'poorly', 'terrible', 'useless', 
        'horrible', 'returning', 
        'flimsy', 'falling apart']
        
        if any(x in text for x in keywords):
            return 1
        else:
            return -1

    # func15: Added more keywords related to 
    # specific product types and issues
    def func15(text):
        keywords = ['pants', 'looks', 
        'toy', 'color', 'camera', 
        'water', 'phone', 'bag', 'worked',
        'arrived', 'lasted', 'fabric', 
        'material', 'build quality', 'finish']
        
        if any(x in text for x in keywords):
            return 1
        else:
            return -1
    \end{python}
    \caption{Refined \DSamazon funcs by LLM (Part 3)}
    \label{fig:amazon_llm_refined_code_3}
\end{figure*}


\begin{tcolorbox}[
    float*=t, 
    width=\textwidth, 
    colback=white, 
    sharp corners, 
    title=LLM LF generation prompt template, 
    label=box:labeling
]

\textbf{Instructions for Generating Labeling Functions}

You are assisting in the generation of labeling functions based on a set of sentences, each provided with its ground truth labels.

\vspace{5pt}
\textbf{Label Information}

Below are the available labels and their corresponding label numbers:
\begin{verbatim}
{class_mapping_string}
\end{verbatim}

\vspace{5pt}
\textbf{Task}

Your objective is to derive labeling functions from the given sentences and their associated labels. The labeling functions should follow one of two formats:
\begin{enumerate}
    \item Keyword-based functions
    \item Regular expression-based functions
\end{enumerate}

Each labeling function should be designed to capture meaningful patterns from the text while ensuring generalizability.

\vspace{5pt}
\textbf{Constraints}
\begin{itemize}
    \item Each labeling function should contain no more than \textbf{10} keywords or regular expressions.
    \item Regular expressions should be used only when a keyword-based function cannot adequately capture a pattern.
    \item Ensure that the selected keywords and patterns are generalizable rather than overly specific.
\end{itemize}

\vspace{5pt}
\textbf{Templates for Labeling Functions}

The two accepted templates for defining labeling functions are shown below:
\begin{itemize}
    \item \textbf{Keyword Template:}
    \begin{verbatim}
    def keyword_[label_name][label_number](x):
        ABSTAIN = -1
         keywords = [list of identified keywords]
        return [label_number] if any(keyword in x for keyword in keywords)
        else ABSTAIN
    \end{verbatim}
    \item \textbf{Regular Expression Template:}
    \begin{verbatim}
    def regex_[label_name][label_number](x):
         ABSTAIN = -1
         return [label_number] if [regular expression related condition]
        else ABSTAIN
    \end{verbatim}
\end{itemize}

\vspace{5pt}
\textbf{Sentence Format}

The sentences are presented in the following format:

\begin{verbatim}
Sentence [Sentence ID]: [Sentence Content]. Label: [Label Text]
\end{verbatim}

Below are the sentences for this task:

\begin{verbatim}
{formatted_sentences}
\end{verbatim}

(where \texttt{formatted\_sentences} is the list of sentences in the specified format)

\vspace{5pt}
\textbf{Final Step}

Using the provided templates, generate appropriate labeling functions based on the given sentences. Ensure that the functions adhere to the constraints and effectively classify the provided text.

\end{tcolorbox}


}


\end{document}